\documentclass{article} % For LaTeX2e

\usepackage[preprint]{neurips_2026}

\usepackage[utf8]{inputenc} % allow utf-8 input
\usepackage[T1]{fontenc}    % use 8-bit T1 fonts
\usepackage{hyperref}       % hyperlinks
\usepackage{url}            % simple URL typesetting
\usepackage{booktabs}       % professional-quality tables
\usepackage{amsfonts}       % blackboard math symbols
\usepackage{nicefrac}       % compact symbols for 1/2, etc.
\usepackage{microtype}      % microtypography
\usepackage[dvipsnames,table]{xcolor}         % colors

\usepackage{multirow}
\usepackage{siunitx}
\usepackage[ruled,vlined]{algorithm2e}
\usepackage{graphicx}

\title{LoRA-generating hypernetworks for efficient on-device LLM generative personalization}

\author{%
 Sean Augenstein\thanks{Corresponding author: \texttt{saugenst@google.com}.}~~\thanks{Google.} \\
 \And
 Li Ding\footnotemark[2] \\
 \And
 Jihwan Lee\thanks{Formerly at Google, work done while at Google.} \\
 \And
 Keith Rush\footnotemark[2] \\
 \And
 Andrey Zhmoginov\footnotemark[3] \\
}

\begin{document}

\maketitle

\begin{abstract}
On-device large language models (`LLMs'), e.g.~running on mobile phones, are ripe for improvement via personalization. The limited compute resources of mobile devices impose limits on model scale and thus model quality, making any realizable quality gains highly impactful. At the same time, their personal nature (i.e., the close coupling to a particular user) means that a given on-device LLM tends to be used in similar, predictable patterns over the course of time. This paper presents a novel method for personalizing on-device LLMs. It trains a \emph{hypernetwork} to map a user's context tokens to a low-rank adaptation (`LoRA') well-suited to that user. Once the trained common artifacts are deployed to users' devices, each user uses the hypernetwork to synthesize (entirely on device) a personalized LoRA. This approach blends the benefits while avoiding the drawbacks of two existing approaches to LLM customization: in-context learning (`ICL') and parameter-efficient fine-tuning (`PEFT'). Like ICL (and unlike PEFT), the on-device phase of our approach is computationally feasible, requiring only forward passes through neural networks. Like PEFT (and unlike ICL), our approach modifies the `target' base LLM via weights (the LoRA), avoiding negative consequences (e.g.~increased latency) associated with extending the input sequence. Our approach is particularly well-suited to the mobile device regime. Apart from the on-device compute and latency benefits mentioned, it also requires minimal additional storage, as internally its architecture partly leverages the same LLM weights as belong to the target LLM to be personalized. We demonstrate the benefits of LoRA-generating hypernetworks on several representative personalization datasets, comparing against baselines like ICL and PEFT. Of note, our personalization experiments focus on more challenging and less studied long-form text generation tasks.

\end{abstract}

\section{Introduction}

As transformer-based large language models (`LLMs') have revolutionized the field of natural language processing (`NLP') in general, so too have they revolutionized mobile device-based NLP. Mobile device ecosystems now include APIs for accessing `on-device' LLMs \citep{android_ai_foundation_2023, apple_intelligence_2024, microsoft_phi_2025} enabling mobile applications to include generative AI features. To fit within the compute- and space-constraints of mobile phones, such on-device LLMs necessarily have fewer parameters and thus are acutely capacity-constrained in the capabilities they can be imbued with during training. This presents a challenge to mobile generative AI developers.

A mobile phone is typically used by only a single user, making a mobile device's usage patterns deeply personal and thus relatively consistent over longer periods of time, i.e., reflecting the device user's usual writing style, tone, etc. This presents an opportunity to mobile generative AI developers; if capable techniques existed for personalizing an on-device LLM to a user's medium-to-long-term usage patterns, this would alleviate the scale-related challenges of mobile LLMs mentioned above.

On-device LLM personalization is an instance of `downstream' LLM customization, whereby a LLM is adapted to a novel task or user without adjusting the LLM’s parameters. There are two predominant approaches. The first is parameter-efficient fine-tuning (`PEFT'), which augments the LLM with additional parameters and trains them such that the augmented LLM performs well for the given task or user. An example in widespread use is low-rank adaptation (`LoRA') \citep{hu2021lora}. The second common approach to LLM customization is in-context learning (`ICL') \citep{brown2020icl}, where the LLM is provided with examples from the given task or user in its input sequence. 

PEFT and ICL each have drawbacks, particularly when applied to on-device LLM personalization. PEFT requires gradient descent to personalize a LoRA, a computationally prohibitive exercise on a mobile device. ICL increases input sequence length and thus inference latency, which is typically undesirable for interactive generative AI features. ICL also suffers worse quality as context length grows\footnotemark{} \citep{liu2023lostinthemiddle, li2024longcontextllmsstruggle}. Considering that an LLM's input sequence will also serve other uses (e.g., as a conversation record during multi-turn chatbot interaction), it may be preferable to utilize alternative manners of personalizing the on-device LLM to its user's long-term characteristics.

\footnotetext{The phrase `context rot' \citep{workaccount22025contextrot} has been coined to describe this degradation as context length grows.}

\begin{table}
    \caption{Comparative strengths of LLM personalization methods. LoRA-generating hypernetworks offer the most attractive cumulative performance on the dimensions that matter the most for on-device LLM-based generative AI, namely, \emph{per-user} compute cost (which will be performed on device) and \emph{per-query} inference latency and quality (both key to a satisfactory user experience). Hypernetworks do involve more up-front training cost (effectively, parameter-efficient backpropagation through two LLMs instead of one, see Section \ref{sec.train}), but this computation is only performed once, off device.}
    \label{tab.llm_personalization_comparison}
        \begin{center}
        \begingroup
        \setlength{\tabcolsep}{4pt}                    
        \begin{tabular}{l|c||c|c|c}
            \toprule
            \multicolumn{1}{c|}{\multirow{5}{*}{method}}
            & compute to & compute to & latency of & quality of \\
            & make common & make personal & inference & inference \\
            & artifacts & artifacts & response & response \\
            & (Phase 1, & (Phase 2, & (Phase 3, & (Phase 3, \\
            & once overall) & once \emph{per-user}) & \emph{per-query}) & \emph{per-query}) \\\hline
            \textsc{(non-personalized)} & \cellcolor{YellowGreen} low & \cellcolor{YellowGreen} N/A & \cellcolor{YellowGreen} lowest & \multirow{6}{*}{\begin{tabular}{l} (see Sec. \ref{sec.exp}) \end{tabular}} \\
            \cline{1-4}
            \textsc{\emph{per-user} ICL} & \cellcolor{YellowGreen} low & \cellcolor{YellowGreen} negligible &  \cellcolor{red} highest & \\
            \cline{1-4}
            \textsc{\emph{per-user} LoRAs} & \cellcolor{YellowGreen} & \cellcolor{red} & \cellcolor{YellowGreen}  \\
            \textsc{(via PEFT)} & \cellcolor{YellowGreen} \multirow{-2}{*}{low} & \cellcolor{red} \multirow{-2}{*}{highest} & \cellcolor{YellowGreen} \multirow{-2}{*}{lowest} & \\
            \cline{1-4}
            \textsc{\emph{per-user} LoRAs} & \cellcolor{red} & \cellcolor{YellowGreen} & \cellcolor{YellowGreen} &  \\
            \textsc{(via Hypernetwork)} & \cellcolor{red} \multirow{-2}{*}{highest} & \cellcolor{YellowGreen} \multirow{-2}{*}{low} & \cellcolor{YellowGreen} \multirow{-2}{*}{lowest} & \\
            \bottomrule
        \end{tabular}
        \endgroup
        \end{center}
    \vskip -0.1in
\end{table}

In this work we present just such an alternative form of personalization, via \emph{LoRA-generating hypernetworks}. Entirely on device, the hypernetwork takes representative examples of a user as input and produces a personalized LoRA as output, which is then attached to the on-device LLM. Like ICL, it has the desirable property that personalization only involves forward passes, so is computationally realizable on device. Like PEFT, the outputs are neural parameters (a LoRA), so we avoid increasing the LLM input sequence, leaving its capacity preserved for other uses. 

Table \ref{tab.llm_personalization_comparison} provides a comparison of various methods of LLM personalization. We distinguish between three phases in a personalization pipeline: common artifact creation (`Phase 1', performed once overall for the population), personal artifact creation (`Phase 2', performed once per-user), and personalized inference (`Phase 3', performed per-query). Given this framing, LoRA-generating hypernetworks are desirable for on-device LLM personalization because they upstream compute cost into Phase 1 (which is the least constrained, as well as amortizable) and out of Phases 2 and 3 (which are the most constrained, as they must take place on device).

Table \ref{tab.llm_personalization_comparison} indicates the advantages of LoRA-generating hypernetworks in \emph{both} per-user computational cost \emph{and} per-query latency; what remains to be determined are the relative quality of the personalized inference responses when comparing to PEFT or ICL. We undertake this comparative evaluation in this paper, showing via experiments on representative generative NLP tasks that hypernetwork-generated LoRAs deliver responses that measure as well or better on quality as the alternatives.

The contributions of this paper are as follows:
\begin{itemize}
    \item The first study (to our knowledge) of LoRA-generating hypernetworks for on-device LLM personalization, considering its advantages over alternatives to the constraints of the mobile device environment, and presenting an architecture tailored to maximize reuse of already-present on-device neural artifacts and allow personalized LoRA generation to take place entirely on-device.
    % \item The first investigation (to our knowledge) of the capabilities of LoRA-generating hypernetworks for more complex \emph{generative} (as opposed to classification) NLP tasks.
    \item Experiments performing personalization of representative on-device LLMs on representative generative tasks, providing empirical evidence that hypernetworks perform equal or better at \emph{quality}, as compared to alternative methods of LLM personalization.    
\end{itemize}

% In particular, we show that LoRA-generating hypernetworks are well-suited for customization of mobile device generative AI applications. In this setting, it is desirable for customization-time processing to take place entirely on-device. Our hypernetwork is LLM-based and shares common parameters with the on-device LLM being adapted, so that synthesizing a LoRA from examples can be performed directly on device.

\paragraph{Some Context on `Context'} NLP personalization can be considered to be of two distinct flavors (both of which may be necessary or helpful to the success of a generative AI product). One flavor is longer-term, persona-related personalization which adapts a model to a user's usage style and traits, as they hold in the medium-to-long term over many queries made to the model (e.g., a user writes to friends in the vernacular of a particular region/language, and writes to work colleagues in formal English). The other flavor is query-related personalization which adapts a model to the immediate needs of the user for a particular query in the moment, often involving retrieval-augmented generation or `RAG' (e.g., the user is writing to a friend to share the details of a particular party invitation they've received). This paper considers the former flavor of personalization, i.e., how to best tailor an on-device LLM to a user's persona and long-term traits. Our position is that LoRA-generating hypernetworks are useful for longer-term personalization, with advantages over PEFT and ICL. We do \emph{not} claim that hypernetworks are indicated for immediate/query-related personalization. Indeed, our belief is that hypernetworks should be adopted for long-term personalization in part so that input sequence capacity can be preserved for use for immediate/query-related personalization.

% Delighting a user may involve addressing both of these aspects in an NLP response to a given query. For example, a user may query an NLP system to ``write an invite to Sarah for my dinner party this weekend, and suggest a side dish she can bring'', and an ideal personalized response would take account of long-term context that holds absent the query (the user's writing style and any food allergies) and immediate context related to the specific query (the data/time/location of the user's party).

\section{Related Work}
\label{sec.rw}

\paragraph{Parameter-Efficient Fine-Tuning (`PEFT')} PEFT freezes a `base' LLM's parameters and augments it with new trainable parameters (much fewer in number than the base). Optimizing the PEFT parameters still requires backpropagation through the base LLM, but parameter storage burdens are greatly reduced. Notable PEFT flavors include adapters \citep{houlsby2019adapters}, prompt tuning \citep{lester2021prompt}, and `LoRA' \citep{hu2021lora}. Personalizing LLMs by per-user PEFT has been studied \citep{collins2023profit, khan2024portllm, tan2025onepeftperuser}, but in general PEFT is computationally infeasible to perform on mobile phones for state-of-the-art on-device LLMs. Such LLMs are sized to just fit inference within a high-end mobile phone's RAM budget\footnotemark{}, and ``[m]emory requirements for fine-tuning ... models are drastically higher than for standard inference'' \citep{gemma4}.
% , to simplify fine-tuning to ad hoc use cases
% low-rank adaptation or 
% (as the documentation of one recently released, mobile-phone-targeted open-source LLM states)

\footnotetext{At time of writing, high-end phones have 12-16 GB of RAM, to support all on-device computation (operating system, applications, and graphics, as well as carveout for the on-device LLM). The weights alone of one representative on-device LLM occupy 3.2 GB (\emph{after} 4-bit quantization)\citep{gemma4}; there are context window-related memory burdens as well.} 

\paragraph{In-context Learning (`ICL')} ICL \citep{brown2020icl, dong2024iclsurvey} involves prepending representative task examples to a query; the LLM then uses this context to formulate a response conditioned to the task. ICL is desirable for task customization as a means of avoiding parameter fine-tuning (and its attendant burdensome requirements for parameter-level model access, training data, and compute resources). However, ICL is not without its own weaknesses. ICL has been shown to struggle qualitatively as context length grows \citep{li2024longcontextllmsstruggle, du2025contextlengthhurts}, and be biased towards examples at the start and end of input sequences \citep{liu2023lostinthemiddle}. As mentioned, ICL's increased input sequence lengths also bring increased latency.

\paragraph{PEFT \emph{vs.}~ICL, PEFT \emph{for} ICL} PEFT and ICL are distinctly different methods of LLM customization, and comparative analyses of the two have been performed, e.g. \citet{liu2022peftbeatsicl, mosbach2023peft_vs_icl}. ICL works well at larger model scales \citep{wei2022emergent} but lags behind fine-tuning approaches at smaller model scales (e.g. 2-10B parameters) \citep{he2025yofo_icl}. To address this, a thread of research has looked at meta-fine-tuning as a means of \emph{enhancing} a model's ICL capability \citep{min2022metaicl, chen2022incontexttuning, he2025yofo_icl}. As we are motivated by on-device settings with smaller scale LLMs, in this paper we represent ICL with such a `PEFT-for-ICL' approach, i.e.~we fine-tune a single LoRA on ICL-prepended examples to achieve best possible ICL performance.

% \paragraph{Hypernetworks} Neural networks which synthesize parameters for other neural networks have long been of interest to the machine learning community, from early research on the notion of `fast' weights and context-dependent weight changes \citep{schmidhuber1992fastweights} to subsequent work focused on recurrent networks (\citet{ha2017}, which coined the phrase `hypernetwork'), to recent work on customizing transformer-based LLMs to ad hoc tasks \citep{phang2023, chen2024generativeadapter, lv2024hyperlora}. 
% Two particular recent works, \citet{charakorn2025texttolora} and \citet{charakorn2026doctolora}, stand out for their successful demonstration of LoRA-generating hypernetwork architectures to representative classification tasks, and for their focus on minimizing the amount of new parameters required in a hypernetwork system. However, LoRA-generating hypernetworks are understudied in on-device personalization applications and in more complex open-ended text generation applications. We focus on both these applications in this paper.

\paragraph{Hypernetworks} \citet{ha2017} coined the phrase `hypernetwork' to refer to neural networks which synthesize parameters of other neural networks, but the concept dates even earlier, to work on `fast' weights and context-dependent weight changes \citep{schmidhuber1992fastweights}. NLP research has studied hypernetworks for customizing LLMs to ad hoc tasks \citep{deb2022boosting, ivison2023hint, phang2023hypertuning, li2024mend, lv2024hyperlora, chen2024generativeadapter}. Some recent works \citep{charakorn2025texttolora, charakorn2026doctolora, liu2026shine} stand out for the quality of their demonstrations of LoRA-generating hypernetworks for realistic tasks. Unfortunately, none of these works address or are directly applicable to the mobile device setting. In a mobile generative AI platform like Android AICore \citep{android_ai_foundation_2023}, the base LLM available on device is an instruction-tuned, decoder-only model and LoRAs are the supported adaptation modality. The previous works listed either consider architectures with models not already present on device, or at parameter scales infeasible to deploy to device, or they don't use LoRA. With the exception of \cite{chen2024generativeadapter}, none consider personalization.

\paragraph{On-Device Personalization and Federated Learning} In our scenario, the data is decentralized over a population of users' mobile devices. Federated learning (`FL') \citep{mcmahan2017fl} studies how to learn over such datasets. FL allows for meta-training a common artifact to be used in on-device personalization (i.e. `Phase 1' of personalization, above). In this vein, \citet{shamsian2021personalized_fl_hyp} proposed FL for hypernetwork training, but didn't consider LLMs. FL has traditionally involved computing gradients on device, which is infeasible for LLMs. A newer variant of FL shifts computations to instead take place in a trusted execution environment (`TEE') at a server \citep{eichner2025tees}, which decouples FL from the computational limitations of mobile devices. The hypernetwork training algorithm presented in this paper is realizable in production via such `TEE-based' FL.
% TODO(saugenst): cite Jakub's intern, who compared FL to MAML?
% FL is meta-training with on-device data, and so naturally relates to on-device model personalization. 

\paragraph{LLM Personalization Datasets} Studying personalization requires a user-partitioned dataset, i.e., a `dataset of per-user datasets' containing a wide spectrum of users, each with examples that evince their particular traits. With the motivation of accelerating research on LLM personalization, several high-quality user-partitioned datasets like LaMP \citep{salemi2024lamp} and LongLaMP \citep{kumar2024longlamp} have been released in recent years. We leverage these public datasets in our experiments.
% The approaches above all involve meta-training a common artifact (e.g.~a LoRA, a hypernetwork, etc.) which induces the model to perform well for different users. In such meta-training, a user-partitioned dataset is used

% \paragraph{User Embeddings} Seeking to avoid the higher latency and computational overhead of input-context-based personalization, \citet{ning2024userembedding} present an approach to LLM personalization where users' distinct properties are represented via embeddings produced by a user encoder (analogous to encoders used for other modalities, like images). The generated user embedding is cross-attended to during LLM inference. We share their motivation, and we also use user embeddings to encode personal traits into a latent space. However, our methods must work with on-device LLMs which were not \emph{a priori} designed for user customization via embedding (i.e., are not equipped with user encoders and are not setup for cross-attention). We instead apply the `LLM2Vec' approach \citep{behnamghader2024llm2vec} for modifying a decoder-only LLM to act as an embedder (to address the lack of user encoder), and our architecture is such that we convert the user embedding to a LoRA via a hypernetwork (thus avoiding cross-attention or other architectural modifications).
% % TODO(saugenst): did the spreadout regularization/FL work (Felix Yu) motivate us to do LLM2Vec?

% Also, talk about clustering to a few LoRAs, and how hypernetworks are effectively a manner of softc-clustering (relates to EigenLoRA interpretation)
\paragraph{LoRAs Reuse; User Embeddings} Some previous work relates to sub-components of our hypernetwork architecture. LoraHub \citep{huang2024lorahub} customizes a LoRA via a combination of pre-existing LoRAs. EigenLoRAx \citep{kaushik2025eigenlorax} goes further and performs an `eigendecomposition' to `principal' LoRAs, for use as building blocks. LoRA-generating hypernetworks (as in this paper) can be considered as a step even further: learning via gradient descent the `principal' LoRAs \emph{along with} a neural mapping from user context to `principal' coefficients. On the other hand, \citet{ning2024userembedding} focus solely on learning this user context neural mapping; the mapping outputs are not LoRA coefficients but rather embeddings to be cross-attended by an encoder-decoder LLM. We share their use of user embeddings to encode personal traits into  a latent space. However, our methods must work with on-device LLMs which are decoder-only and not setup for cross-attention, so we instead convert the user embedding to a LoRA inside the hypernetwork (as described next).

\section{Hypernetwork Architecture}
\label{sec.arch}

Our goal is a personalization approach that can be performed entirely within the compute and storage of a mobile device (apart from Phase 1, which only happens once overall for the user population). As such, our hypernetwork involves minimal new parameters and maximal reuse of pretrained artifacts already present on the device. We present the architecture and how it is used for personal artifact creation and inference (Phases 2 and 3), and then present how it is trained (Phase 1) in Section \ref{sec.train}.

The inputs to the hypernetwork are a group of examples providing context about a task and a user's desired manner of satisfying that task, i.e., each example is a pair of query/desired response sequences. The hypernetwork can take this set of example sequences either as a batch (i.e., fed and processed in parallel) or as a single concatenated sequence (like ICL); we find the latter to be more effective, and use that approach in all experiments presented here. The output of the hypernetwork is a set of LoRA matrices. That is, for every weight matrix in the `target' LLM that should be modified with a low-rank adaptation, the hypernetwork produces an `A' and `B' pair of low-rank adaptation matrices. Put together, the LoRA-generating hypernetwork can be considered a form of personalization which resembles ICL on the input side, in that the context is fed as input tokens to a neural network, and resembles PEFT on the output side, in that the product is a set of LoRA matrices.

\begin{figure}
    \centering
    \includegraphics[trim={0cm 14cm 0cm 7cm},clip,width=1.0\textwidth]{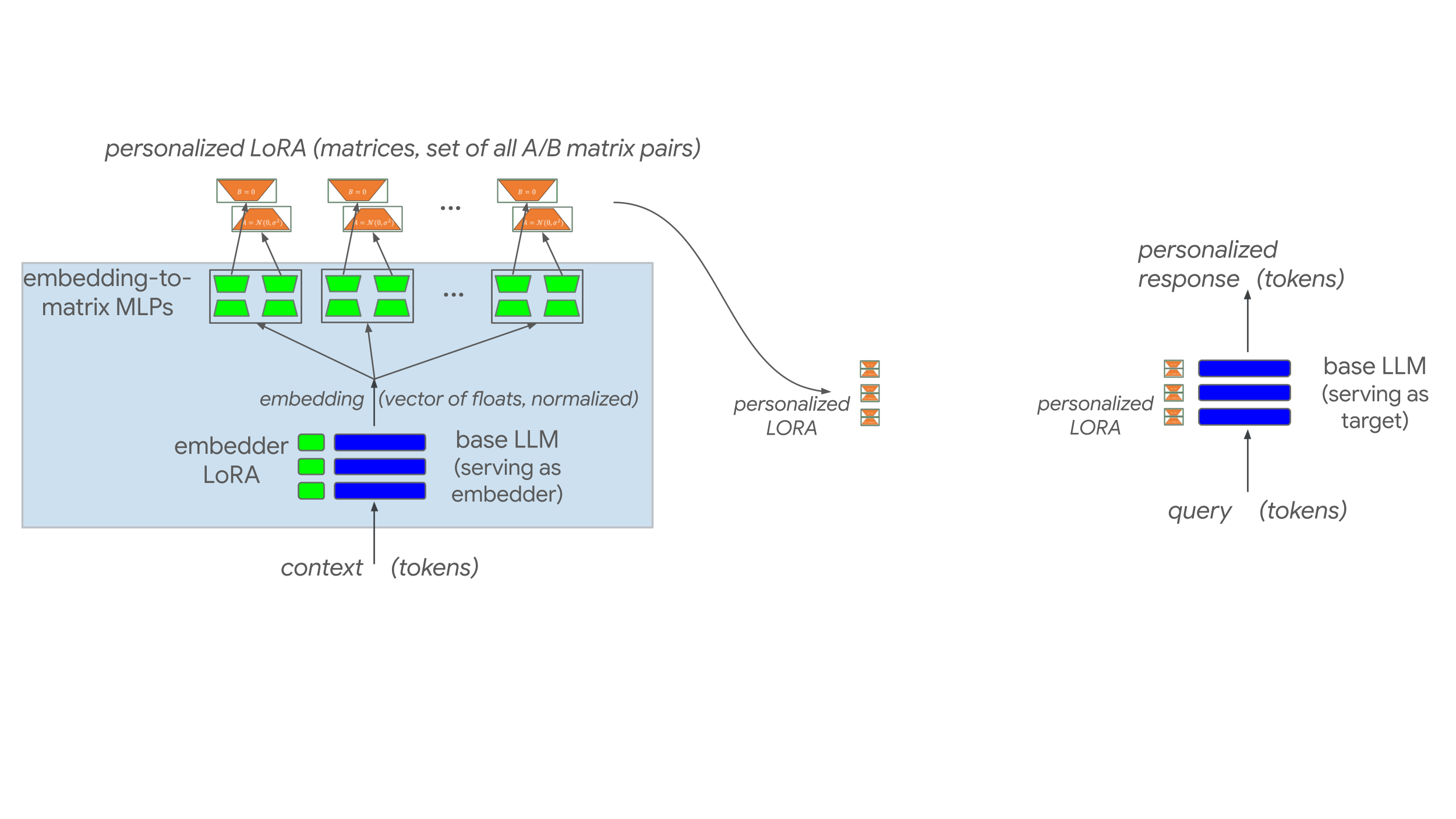}
    \caption{Hypernetwork personal artifact creation (\emph{left}) and personalized inference (\emph{right}), Phases 2 and 3 of the personalization pipeline (respectively). Both take place \emph{entirely on device}. On a given user's device, Phase 2 is only performed once (or infrequently), to generate and store the personal artifacts (a personalized LoRA). Afterwards, anytime a query is issued by the user (Phase 3), the personalized LoRA is attached to the base LLM to induce a personalized response.}
    \label{fig.hypernetwork_p13n_and_inference}
\end{figure}

There is already a trained, frozen LLM located on device: the `target' LLM that is the object of personalization. To save the need for additional parameters, we make use of this LLM in the hypernetwork. It serves as the base of an embedder which maps user context tokens to an embedding vector (encoding the user in a latent user space). The task and the population of users require a custom latent space, and since the on-device LLM cannot be modified, we attach a learnable `embedder LoRA' to the frozen base LLM to convert the LLM into a bespoke text encoder (as in \citet{behnamghader2024llm2vec}). The rank of the embedder LoRA ($r$) is a hyperparameter. The pooling method for aggregating the embedding vector from the LLM's output sequence is discussed in Appendix \ref{app.hypernetwork}.

The second part of the hypernetwork are the `embedding-to-matrix' MLPs, a bank of parallel 2-layer feed-forward blocks. Each 2-layer MLP takes the embedding vector as input and produces as output the weights of a single `A' or `B' LoRA matrix for a particular layer and module of the target LLM. In totality an entire set of LoRAs is produced, augmenting all relevant weight matrices in the target LLM. See Figure \ref{fig.hypernetwork_p13n_and_inference}, \emph{left}.

The 2-layer feed-forward blocks are `bottlenecked', i.e., the intermediate vector between the layers is of much smaller dimension than the embedding vector (the input to the feed-forward blocks) or than an `A' or `B' LoRA matrix (the output of a feed-forward block). One interpretation, which aligns with EigenLoRAx \citep{kaushik2025eigenlorax}, is that the intermediate vector is a vector of coefficients, with the second (or `top') layer of the 2-layer feed-forward block serving as a learned set of `principal' or `eigen'-low-rank-matrices. The first (or `bottom') layer of the 2-layer feed-forward block is a mapping from a user's representation to the corresponding coefficients for the `principal' matrices which create the best LoRA matrix for the user in question. The width of the bottleneck is a hyperparameter ($k$), and selection of its value can be interpreted as roughly analogous to selecting the number of top values to retain in a low-rank singular value decomposition problem.

Table \ref{tab.parameter_counts} gives representative parameter counts, for the base LLM and other parts of the hypernetwork. Once a LoRA is generated, the hypernetwork additional parameters can be deleted from device.

\section{Hypernetwork Training}
\label{sec.train}

\begin{figure}
    \centering
     % {left bottom right top}
    \includegraphics[trim={0cm 3cm 0cm 7cm},clip,width=1.0\textwidth]{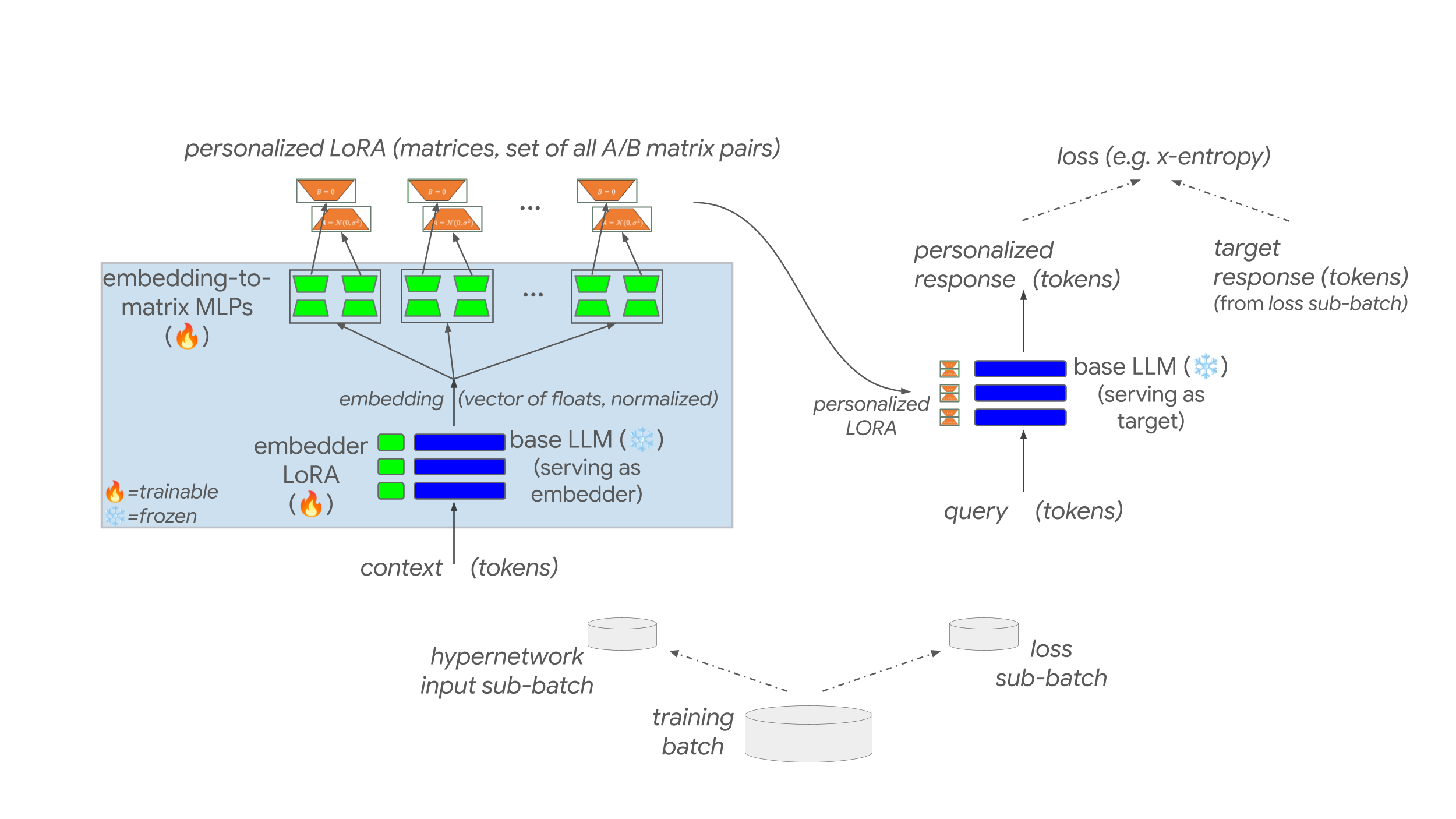}
    \caption{Hypernetwork common artifact creation, Phase 1 of the personalization pipeline. Performed once overall, and then the common artifacts generated (the weights of the embedder LoRA and embedding-to-matrix MLPs, in green) are distributed to all devices in the inference population.}
    \label{fig.hypernetwork_training}
\end{figure}

The hypernetwork is trained in two phases; a pretraining of just the embedder LoRA parameters of the hypernetwork, and then an end-to-end training updating all trainable parameters of the hypernetwork.

\paragraph{LLM2Vec-style Pretraining} The embedder LoRA is pretrained via unsupervised contrastive learning. This teaches the embedder portion of the hypernetwork to distinguish users based on language characteristics present in their context examples, so that distinct users produce distinct embedding vectors that are distant from each other in the latent user space. The technique used is a straightforward application of the contrastive learning step of LLM2Vec (\citet{behnamghader2024llm2vec}, which is itself inspired by the SimCSE algorithm, \citet{gao2022simcse}).

\paragraph{End-to-end Training} Following pretraining, we train all trainable parameters (both the embedder LoRA and the embedding-to-matrix MLPs) via an end-to-end loss. See Figure \ref{fig.hypernetwork_training} and Algorithm \ref{algo.hyp_train}. At each step of training, a cohort of users is sampled. For each user, two disjoint batches of data are formed, one a batch of contextual examples to be used as the hypernetwork's input and the other a batch to be used as the target LLM's input and desired output. For each user, a LoRA is generated (with the former batch) and attached to the target LLM, and then a token cross-entropy loss is calculated with the predictions of this augmented target LLM (using the latter batch). Loss is minimized with respect to the hypernetwork's trainable parameters. In this way, parameters are learned which produce the best LoRAs for a wide variety of users. While training is more computationally intensive in that it involves backpropagation through two LLMs (the target LLM and the hypernetwork) instead of one, recall that this phase of the personalization pipeline takes place off device and only happens once. See Appendix \ref{app.hypernetwork} for further details on hypernetwork training.

\paragraph{Distillation} An additional aspect that proved highly beneficial at end-to-end training time: using ICL as a teacher and distilling it into the hypernetwork. In this manner we are effectively performing context distillation \citep{snell2022contextdistillation}, except instead of a fixed context, we continually show the ICL and the student hypernetwork different (users') contexts, so that the hypernetwork learns to personalize to various contexts following the soft labels of the teacher ICL. We explore replacing this distillation with an alternative approach in Appendix \ref{app.ablations}.

\section{Experiments}
\label{sec.exp}

We measure the capabilities of LoRA-generating hypernetworks on several representative personalization tasks, with several different model configurations, and with comparisons to alternative methods of LLM personalization.

\subsection{Personalization Tasks (Datasets)}
\label{subsec.datasets}

We consider three representative personalization tasks, selected because they are challenging \emph{generation} tasks which have been understudied for personalization \citep{kumar2024longlamp}, and (to our knowledge) understudied via hypernetwork-based approaches. The tasks are described next, with additional details in Appendix \ref{app.datasets}. Each is a `dataset of datasets', consisting of many users (with varied traits) each having multiple examples (reflecting the particular user's traits). In our experiments we use disjoint sets of users for train and test, to gauge generalization of personalization ability. 

\paragraph{Personalized Amazon Review Writing (LongLaMP-3)} LongLaMP \cite{kumar2024longlamp} introduced several challenging long-text generation tasks, with the objective of seeing them studied for personalization. LongLaMP-3 is a dataset of Amazon reviews \citep{ni2019amazon}; each example consists of a product summary (written by Amazon) and a description, score, and in-depth review (written by an Amazon user) for that product. The task is to generate a user-personalized in-depth review, given the other information.

\paragraph{Personalized Reddit Post Writing (LongLaMP-4)} In this dataset (based on \citet{vlske2017reddit}), the users are Reddit contributors, and each example is content from a Reddit post along with a summary of the post. The task is to generate content for a post that is personalized (i.e., reflecting the user's writing style and general interests), given the post's summary. 

\paragraph{Personalized Scholarly Title Writing (LaMP-5)} This is from an earlier set of LaMP datasets \citep{salemi2024lamp}, and involves shorter sequence generation than the aforementioned LongLaMP datasets. In this dataset the users are researchers and the examples are scholarly articles and their titles (based on the Citation Network Dataset of \cite{tang2008citation}). The task is to generate the title for a given scholarly article, personalized to reflect the researcher's academic interests.

\subsection{Comparison Baselines}
\label{subsec.baselines}

We compare LoRA-generating hypernetworks to a number of relevant baselines. The common aspect to all the approaches considered is that they never modify the base LLM's weights; they only learn parameter-efficient adaptations to the weights (e.g.~LoRAs) or adjust LLM inputs (e.g.~ICL prefixes). 
% In this way, they can all be considered valid manners of `downstream' LLM customization (as discussed in the Introduction).

% \paragraph{Unmodified base LLM} The simplest baseline; a `stock' on-device scale LLM. Non-personalized.

% \paragraph{Unmodified base LLM with per-user ICL} The simplest personalized baseline. The LLM is evaluated on examples modified with user-specific ICL prefixes, to personalize behavior accordingly.

\paragraph{Single common LoRA (\textsc{`Non-personal baseline'})} A single LoRA is trained across all the users in the training dataset, and evaluated on all users in the test dataset, to measure the best non-personalized baseline. The expectation is that a useful personalization method should in general exceed this baseline.

\paragraph{Single common LoRA, tuned for per-user ICL (\textsc{`ICL'})} Like above, a single common LoRA is learned; however, at training time it is shown examples with user-specific ICL prefixes attached, so that it `learns' to customize to user via ICL. At test time, the LLM is fed ICL-modified input examples, to personalize behavior accordingly. This is essentially the approach of MetaICL \citep{min2022metaicl}, except here we only fine-tune via LoRA (we don't fine-tune the base LLM's weights).

\paragraph{Per-user LoRAs via PEFT (\textsc{`PEFT'})} Each user trains its own individual LoRA via gradient descent, as in \citet{tan2025onepeftperuser, khan2024portllm}. The gradient descent is initialized from the single common LoRA mentioned above. The optimization is performed with Adam \citep{kingma2017adam}, with all users using a common learning rate (selected as best via hyperparameter sweep). % More details on per-user LoRA fine-tuning in Appendix \ref{app.}.

\paragraph{Per-user LoRAs via hypernetwork (\textsc{`Hypernetwork'})} The method of Sections \ref{sec.arch} and \ref{sec.train}.

Hyperparameter settings and other details are presented in Appendix \ref{app.hyperparameters}.

\subsection{Evaluation Metrics}
\label{subsec.metrics}

As our focus is on text generation tasks, we follow previous works \citep{salemi2024lamp, kumar2024longlamp} and use ROUGE scores \citep{lin2004rouge} for evaluating quality of generated text. Calculating ROUGE score (or any other generative metric) involves the added complexity of sampling/decoding from the personalized LLM to generate full prediction sequences. The decoding process involves a temperature $T$ hyperparameter, affecting the `novelty' of sequences produced. ROUGE score varies significantly with $T$. Consequently, for each personalization method and dataset, we performed a sweep over $T$ to determine the best ROUGE score in that scenario. We determined that a $T=1.0$ resulted in best ROUGE scores for LongLaMP-3 and LongLaMP-4 (for all personalization methods), and $T=0.0$ (i.e.~`greedy' decoding) worked best for LaMP-5.

At temperatures greater than 0, for a given query the ROUGE score can vary significantly from one prediction to the next. To reduce noise in measurement, for LongLaMP-3 and LongLaMP-4 (where $T=1.0$), in every test scenario, for every individual example query in the test set, we generated 10 predictions and computed the average ROUGE score over the predictions. This gave us a lower variance, tighter measure of a given user's ROUGE score (under a particular scenario).

To assess personalization quality, we considered ROUGE score in two ways over the population of users.  The first was the ROUGE score averaged over users. The second is the percentage of users who see an improvement in ROUGE score when comparing against what they'd experience with a single common LoRA (the best non-personal baseline). Both are important when evaluating personalization approaches; the mean conveys improvement \emph{depth} and the improvement percentage conveys improvement \emph{breadth}.

\subsection{Setup}
\label{subsec.setup}

We evaluate generative personalization performance with two different base LLMs and two different LoRA ranks. We use \textsc{Gemma3-1B-IT} \citep{gemma3} and \textsc{Gemma1-2B-IT} \citep{gemma}, both `smaller' LLMs at scales representative of those used on mobile devices. As a means of comparing things as equivalently as possible, we compare personalization methods where the LoRA used at personalized inference time is of identical rank. E.g., we compare the non-personal baseline LoRA at rank=4, the ICL-trained LoRA at rank=4, and have PEFT and the hypernetwork both creating per-user LoRAs that are of rank=4. In our experiments, we used `attention LoRA', i.e., we generated and applied low-rank adaptations for only the attention-related matrices of the base LLM. This was merely an implementation choice for experimentation (hypernetworks are equally capable of generating LoRAs that modify feedforward-related matrices).

\subsection{Results}
\label{subsec.results}

\begin{figure}
    \centering
    \includegraphics[trim={0cm 0cm 0cm 0cm},clip,width=1.0\textwidth]{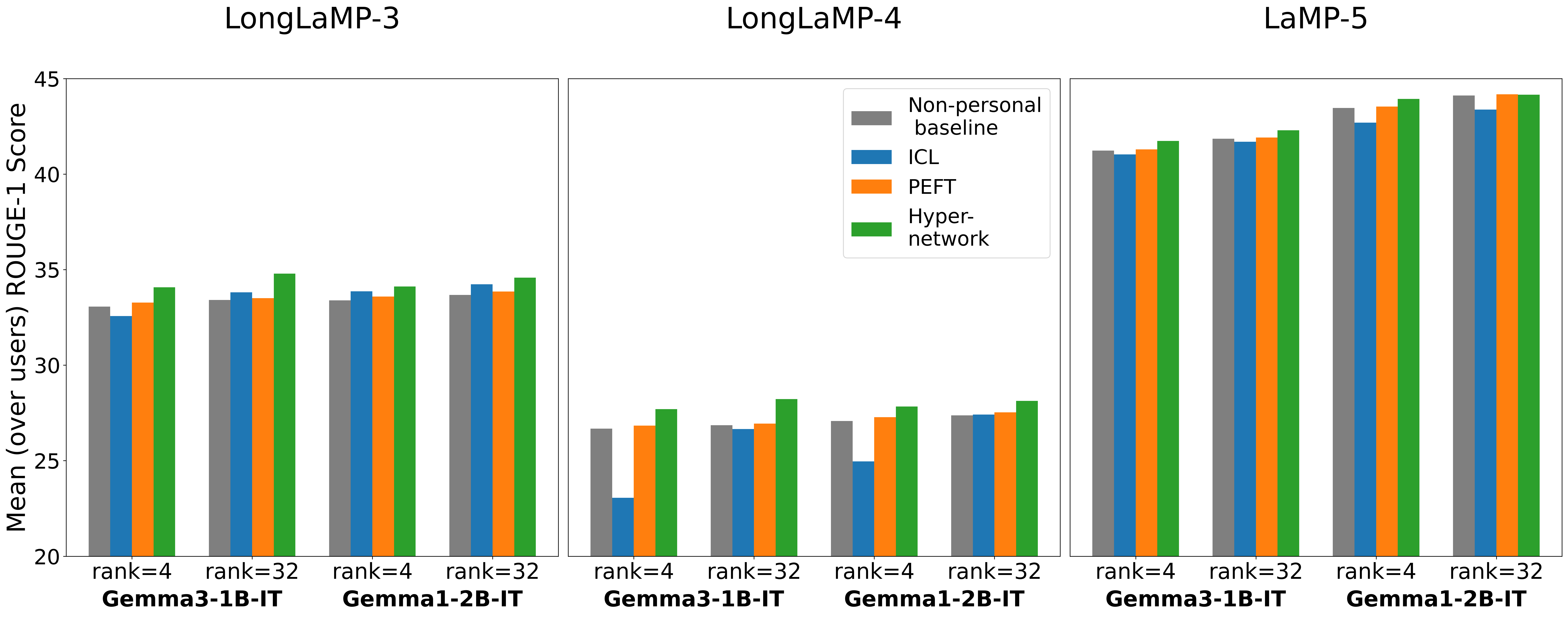}
    \caption{Average (over users) ROUGE-1 score for different personalization methods (ICL/PEFT/Hypernetwork), under various personalizations scenarios (datasets $\times$ base LLM models $\times$ rank of modifying LoRA). Hypernetworks consistently achieve  highest average ROUGE-1.}
    \label{fig.mean_rouge}
\end{figure}

\begin{figure}
    \centering
    \includegraphics[trim={0cm 0cm 0cm 0cm},clip,width=1.0\textwidth]{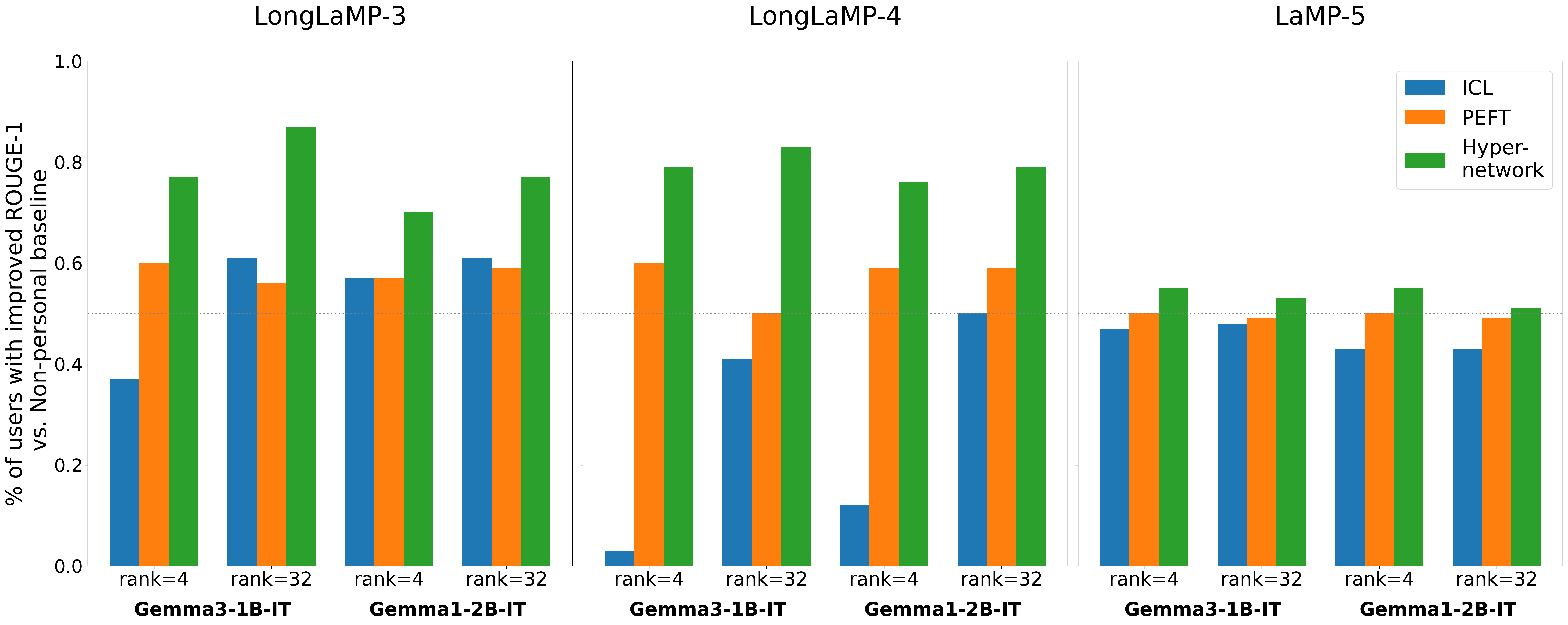}
    \caption{\% of users who's personal ROUGE-1 score improves if switching from the best non-personalized baseline to a given personalization method (ICL/PEFT/Hypernetwork). Hypernetworks consistently improve ROUGE-1 scores for the most users.}
    \label{fig.improvement_percentage}
\end{figure}

Figures \ref{fig.mean_rouge} and \ref{fig.improvement_percentage} summarize the results of personalization experiments on the LongLaMP-3, LongLaMP-4, and LaMP-5 datasets. The hypernetwork-generated LoRAs exhibit the best ROUGE-1 score performance (or nearly so) in all scenarios. With LongLaMP-3 and LongLaMP-4, for almost all the configurations, hypernetworks improved the ROUGE-1 score for more than three quarters of users. The LaMP-5 dataset (Table \ref{tab.lamp5_exps_summary}) is challenging to personalize, not just for hypernetworks but for all the other methods as well.

For a more detailed breakdown of results, see Tables \ref{tab.ll3_exps_summary}, \ref{tab.ll4_exps_summary}, and \ref{tab.lamp5_exps_summary} in Appendix \ref{app.results}. We also studied a few ablations, namely, changes in how the hypernetwork is changed, and swapping of context examples to validate that the hypernetwork is actually using context information. See Appendix \ref{app.ablations}.

Interestingly, while ICL did the best job of minimizing cross-entropy (shown in Appendix \ref{app.results}), it did the poorest job at maximizing ROUGE-1 score. This could in part be indicative that the predictive distribution of ICL is more peaked than the hypernetworks, but increasing the decoding temperature beyond $T=1.0$ (which should further flatten out this predictive distribution) did not result in any further increase in ROUGE-1 score.

\section{Conclusions and Future Work}

This paper has presented the concept of using LoRA-generating hypernetworks for on-device LLM personalization, described an architecture that leverages already-present on-device networks, and demonstrated performance of the hypernetwork that equals or surpasses other personalization approaches, under several challenging generative tasks.
 
We chose what we believe are a challenging set of personalization datasets. While we observe positive results with LoRA-generating hypernetworks, further study would aid in determining if there are particular types of personalization scenarios where hypernetworks perform poorly for some users. For example, hypernetworks (like ICL) rely in some sense on `wisdom of the crowd', that a particular user (at test time, i.e. Phases 2 and 3 in Table \ref{tab.llm_personalization_comparison}) has traits that are similar to (or an interpolation between) users that contributed at training time (i.e., Phase 1). An iconoclastic test-time user would be underserved by hypernetwork-based personalization (or ICL-based personalization, for that matter). For such an `out-of-distribution' user, only a parameter fine-tuning-based approach would be reasonably expected to yield significant quality improvement.

In the same vein, production usage of hypernetworks should involve the data of actual users at hypernetwork training time (Phase 1), to minimize chances that a test-time user will be `out-of-distribution'. The participation of these real users would need to include privacy protections. As mentioned in Section \ref{sec.rw}, the Phase 1 training of a personalizing hypernetwork is performed via federated learning (`FL') \citep{mcmahan2017fl} leveraging trusted execution environments (`TEEs') \cite{eichner2025tees}. FL can be composed with differential privacy (`DP') \citep{mcmahan2018dpfl} to provide anonymization to training-time participants, but the inclusion of DP brings additional complexities (e.g., hyperparameter selection to balance useful quality with meaningful privacy). The study of hypernetwork training under DP would be a meaningful next step in advancing hypernetwork-based personalization to production readiness.

The potentialities of hypernetworks for on-device computing are exciting, as they portend a ``cognitive core" future \citep{karpathy2025cognitivecoretweet} where base LLMs are stripped down to minimal size necessary to focus exclusively on general `capability', with user/task-specific facets, e.g. useful `encyclopedic knowledge' or stylistic preferences, handled via cloud look-ups (if short-term/query-specific) or via swappable modular hypernetwork-generated LoRAs (if holding over many queries). We are excited by this vision, and we fervently encourage additional research on hypernetworks for mobile device computing in order to make it a reality.

\subsubsection*{Acknowledgments}

The authors thank Zachary Garrett, Brendan McMahan, and Zachary Charles for generous support and feedback at various stages of the research presented here.

\bibliography{iclr2025_conference}

@misc{android_ai_foundation_2023,
  author       = {{Android Developers}},
  title        = {A New Foundation for {AI} on {Android}},
  howpublished = {Android Developers Blog},
  year         = {2023},
  month        = {December},
  day          = {6},
  url          = {https://android-developers.googleblog.com/2023/12/a-new-foundation-for-ai-on-android.html},
  note         = {Accessed: 2026-04-21}
}

@misc{apple_intelligence_2024,
  author       = {{Apple Inc.}},
  title        = {Introducing {Apple} Intelligence for {iPhone}, {iPad}, and {Mac}},
  howpublished = {Apple Newsroom},
  year         = {2024},
  month        = {June},
  day          = {10},
  url          = {https://www.apple.com/newsroom/2024/06/introducing-apple-intelligence-for-iphone-ipad-and-mac/},
  note         = {Accessed: 2026-04-21}
}

@misc{behnamghader2024llm2vec,
      title={LLM2Vec: Large Language Models Are Secretly Powerful Text Encoders}, 
      author={Parishad BehnamGhader and Vaibhav Adlakha and Marius Mosbach and Dzmitry Bahdanau and Nicolas Chapados and Siva Reddy},
      year={2024},
      eprint={2404.05961},
      archivePrefix={arXiv},
      primaryClass={cs.CL},
      url={https://arxiv.org/abs/2404.05961}, 
}

@misc{brown2020icl,
      title={Language Models are Few-Shot Learners}, 
      author={Tom B. Brown and Benjamin Mann and Nick Ryder and Melanie Subbiah and Jared Kaplan and Prafulla Dhariwal and Arvind Neelakantan and Pranav Shyam and Girish Sastry and Amanda Askell and Sandhini Agarwal and Ariel Herbert-Voss and Gretchen Krueger and Tom Henighan and Rewon Child and Aditya Ramesh and Daniel M. Ziegler and Jeffrey Wu and Clemens Winter and Christopher Hesse and Mark Chen and Eric Sigler and Mateusz Litwin and Scott Gray and Benjamin Chess and Jack Clark and Christopher Berner and Sam McCandlish and Alec Radford and Ilya Sutskever and Dario Amodei},
      year={2020},
      eprint={2005.14165},
      archivePrefix={arXiv},
      primaryClass={cs.CL},
      url={https://arxiv.org/abs/2005.14165}, 
}

@misc{charakorn2025texttolora,
      title={Text-to-LoRA: Instant Transformer Adaption}, 
      author={Rujikorn Charakorn and Edoardo Cetin and Yujin Tang and Robert Tjarko Lange},
      year={2025},
      eprint={2506.06105},
      archivePrefix={arXiv},
      primaryClass={cs.LG},
      url={https://arxiv.org/abs/2506.06105}, 
}

@misc{charakorn2026doctolora,
      title={Doc-to-LoRA: Learning to Instantly Internalize Contexts}, 
      author={Rujikorn Charakorn and Edoardo Cetin and Shinnosuke Uesaka and Robert Tjarko Lange},
      year={2026},
      eprint={2602.15902},
      archivePrefix={arXiv},
      primaryClass={cs.CL},
      url={https://arxiv.org/abs/2602.15902}, 
}

@misc{chen2022incontexttuning,
      title={Meta-learning via Language Model In-context Tuning}, 
      author={Yanda Chen and Ruiqi Zhong and Sheng Zha and George Karypis and He He},
      year={2022},
      eprint={2110.07814},
      archivePrefix={arXiv},
      primaryClass={cs.CL},
      url={https://arxiv.org/abs/2110.07814}, 
}

@misc{chen2024generativeadapter,
      title={Generative Adapter: Contextualizing Language Models in Parameters with A Single Forward Pass}, 
      author={Tong Chen and Hao Fang and Patrick Xia and Xiaodong Liu and Benjamin Van Durme and Luke Zettlemoyer and Jianfeng Gao and Hao Cheng},
      year={2024},
      eprint={2411.05877},
      archivePrefix={arXiv},
      primaryClass={cs.LG},
      url={https://arxiv.org/abs/2411.05877}, 
}

@misc{collins2023profit,
      title={Profit: Benchmarking Personalization and Robustness Trade-off in Federated Prompt Tuning}, 
      author={Liam Collins and Shanshan Wu and Sewoong Oh and Khe Chai Sim},
      year={2023},
      eprint={2310.04627},
      archivePrefix={arXiv},
      primaryClass={cs.LG},
      url={https://arxiv.org/abs/2310.04627}, 
}

@misc{deb2022boosting,
      title={Boosting Natural Language Generation from Instructions with Meta-Learning}, 
      author={Budhaditya Deb and Guoqing Zheng and Ahmed Hassan Awadallah},
      year={2022},
      eprint={2210.11617},
      archivePrefix={arXiv},
      primaryClass={cs.CL},
      url={https://arxiv.org/abs/2210.11617}, 
}

@misc{dong2024iclsurvey,
      title={A Survey on In-context Learning}, 
      author={Qingxiu Dong and Lei Li and Damai Dai and Ce Zheng and Jingyuan Ma and Rui Li and Heming Xia and Jingjing Xu and Zhiyong Wu and Tianyu Liu and Baobao Chang and Xu Sun and Lei Li and Zhifang Sui},
      year={2024},
      eprint={2301.00234},
      archivePrefix={arXiv},
      primaryClass={cs.CL},
      url={https://arxiv.org/abs/2301.00234}, 
}

@misc{du2025contextlengthhurts,
      title={Context Length Alone Hurts LLM Performance Despite Perfect Retrieval}, 
      author={Yufeng Du and Minyang Tian and Srikanth Ronanki and Subendhu Rongali and Sravan Bodapati and Aram Galstyan and Azton Wells and Roy Schwartz and Eliu A Huerta and Hao Peng},
      year={2025},
      eprint={2510.05381},
      archivePrefix={arXiv},
      primaryClass={cs.CL},
      url={https://arxiv.org/abs/2510.05381}, 
}

@misc{eichner2025tees,
      title={Confidential Federated Computations}, 
      author={Hubert Eichner and Daniel Ramage and Kallista Bonawitz and Dzmitry Huba and Tiziano Santoro and Brett McLarnon and Timon Van Overveldt and Nova Fallen and Peter Kairouz and Albert Cheu and Katharine Daly and Adria Gascon and Marco Gruteser and Brendan McMahan},
      year={2025},
      eprint={2404.10764},
      archivePrefix={arXiv},
      primaryClass={cs.CR},
      url={https://arxiv.org/abs/2404.10764}, 
}

@misc{gao2022simcse,
      title={SimCSE: Simple Contrastive Learning of Sentence Embeddings}, 
      author={Tianyu Gao and Xingcheng Yao and Danqi Chen},
      year={2022},
      eprint={2104.08821},
      archivePrefix={arXiv},
      primaryClass={cs.CL},
      url={https://arxiv.org/abs/2104.08821}, 
}

@misc{gemma,
      title={Gemma},
      url={https://arxiv.org/abs/2403.08295},
      publisher={Google DeepMind},
      author={{Gemma Team}},
      year={2024}
}

@misc{gemma3,
    title={Gemma 3},
    url={https://arxiv.org/abs/2503.19786},
    publisher={Google DeepMind},
    author={{Gemma Team}},
    year={2025}
}

@misc{gemma4,
      author       = {{Google Inc.}},
      title        = {Gemma 4 model overview},
      howpublished = {Google AI for Developers},
      year         = {2026},
      month        = {May},
      day          = {5},
      url          = {https://ai.google.dev/gemma/docs/core},
      note         = {Accessed: 2026-05-06}
}

@inproceedings{ha2017,
  author       = {David Ha and
                  Andrew M. Dai and
                  Quoc V. Le},
  title        = {HyperNetworks},
  booktitle    = {5th International Conference on Learning Representations, {ICLR} 2017,
                  Toulon, France, April 24-26, 2017, Conference Track Proceedings},
  _publisher    = {OpenReview.net},
  year         = {2017},
  _url          = {https://openreview.net/forum?id=rkpACe1lx},
}

@misc{he2025yofo_icl,
      title={You Only Fine-tune Once: Many-Shot In-Context Fine-Tuning for Large Language Models}, 
      author={Wenchong He and Liqian Peng and Zhe Jiang and Alex Go},
      year={2026},
      eprint={2506.11103},
      archivePrefix={arXiv},
      primaryClass={cs.CL},
      url={https://arxiv.org/abs/2506.11103}, 
}

@misc{houlsby2019adapters,
      title={Parameter-Efficient Transfer Learning for NLP}, 
      author={Neil Houlsby and Andrei Giurgiu and Stanislaw Jastrzebski and Bruna Morrone and Quentin de Laroussilhe and Andrea Gesmundo and Mona Attariyan and Sylvain Gelly},
      year={2019},
      eprint={1902.00751},
      archivePrefix={arXiv},
      primaryClass={cs.LG}
}

@misc{hu2021lora,
      title={LoRA: Low-Rank Adaptation of Large Language Models}, 
      author={Edward J. Hu and Yelong Shen and Phillip Wallis and Zeyuan Allen-Zhu and Yuanzhi Li and Shean Wang and Lu Wang and Weizhu Chen},
      year={2021},
      eprint={2106.09685},
      archivePrefix={arXiv},
      primaryClass={cs.CL}
}

@misc{huang2024lorahub,
      title={LoraHub: Efficient Cross-Task Generalization via Dynamic LoRA Composition}, 
      author={Chengsong Huang and Qian Liu and Bill Yuchen Lin and Tianyu Pang and Chao Du and Min Lin},
      year={2024},
      eprint={2307.13269},
      archivePrefix={arXiv},
      primaryClass={cs.CL},
      url={https://arxiv.org/abs/2307.13269}, 
}

@inproceedings{ivison2023hint,
    title = "{HINT}: Hypernetwork Instruction Tuning for Efficient Zero- and Few-Shot Generalisation",
    author = "Ivison, Hamish  and
      Bhagia, Akshita  and
      Wang, Yizhong  and
      Hajishirzi, Hannaneh  and
      Peters, Matthew",
    editor = "Rogers, Anna  and
      Boyd-Graber, Jordan  and
      Okazaki, Naoaki",
    booktitle = "Proceedings of the 61st Annual Meeting of the Association for Computational Linguistics (Volume 1: Long Papers)",
    month = jul,
    year = "2023",
    address = "Toronto, Canada",
    publisher = "Association for Computational Linguistics",
    url = "https://aclanthology.org/2023.acl-long.631/",
    doi = "10.18653/v1/2023.acl-long.631",
    pages = "11272--11288"
}

@misc{karpathy2025cognitivecoretweet,
    author = {Andrej Karpathy},
    title = {The race for LLM "cognitive core" - a few billion param model that maximally sacrifices encyclopedic knowledge for capability. {I}t lives... [{X Post}]},
    year = {2025},
    month = {Jun},
    day = {27},
    url = {https://x.com/karpathy/status/1938626382248149433},
}

@misc{kaushik2025eigenlorax,
      title={EigenLoRAx: Recycling Adapters to Find Principal Subspaces for Resource-Efficient Adaptation and Inference}, 
      author={Prakhar Kaushik and Ankit Vaidya and Shravan Chaudhari and Alan Yuille},
      year={2025},
      eprint={2502.04700},
      archivePrefix={arXiv},
      primaryClass={cs.LG},
      url={https://arxiv.org/abs/2502.04700}, 
}

@misc{khan2024portllm,
      title={PortLLM: Personalizing Evolving Large Language Models with Training-Free and Portable Model Patches}, 
      author={Rana Muhammad Shahroz Khan and Pingzhi Li and Sukwon Yun and Zhenyu Wang and Shahriar Nirjon and Chau-Wai Wong and Tianlong Chen},
      year={2024},
      eprint={2410.10870},
      archivePrefix={arXiv},
      primaryClass={cs.CL},
      url={https://arxiv.org/abs/2410.10870}, 
}

@misc{kingma2017adam,
      title={Adam: A Method for Stochastic Optimization}, 
      author={Diederik P. Kingma and Jimmy Ba},
      year={2017},
      eprint={1412.6980},
      archivePrefix={arXiv},
      primaryClass={cs.LG},
      url={https://arxiv.org/abs/1412.6980}, 
}

@misc{kumar2024longlamp,
      title={LongLaMP: A Benchmark for Personalized Long-form Text Generation}, 
      author={Ishita Kumar and Snigdha Viswanathan and Sushrita Yerra and Alireza Salemi and Ryan A. Rossi and Franck Dernoncourt and Hanieh Deilamsalehy and Xiang Chen and Ruiyi Zhang and Shubham Agarwal and Nedim Lipka and Chien Van Nguyen and Thien Huu Nguyen and Hamed Zamani},
      year={2024},
      eprint={2407.11016},
      archivePrefix={arXiv},
      primaryClass={cs.CL},
      url={https://arxiv.org/abs/2407.11016}, 
}

@misc{lester2021prompt,
      title={The Power of Scale for Parameter-Efficient Prompt Tuning}, 
      author={Brian Lester and Rami Al-Rfou and Noah Constant},
      year={2021},
      eprint={2104.08691},
      archivePrefix={arXiv},
      primaryClass={cs.CL}
}

@misc{li2024longcontextllmsstruggle,
      title={Long-context LLMs Struggle with Long In-context Learning}, 
      author={Tianle Li and Ge Zhang and Quy Duc Do and Xiang Yue and Wenhu Chen},
      year={2024},
      eprint={2404.02060},
      archivePrefix={arXiv},
      primaryClass={cs.CL},
      url={https://arxiv.org/abs/2404.02060}, 
}

@misc{li2024mend,
      title={MEND: Meta dEmonstratioN Distillation for Efficient and Effective In-Context Learning}, 
      author={Yichuan Li and Xiyao Ma and Sixing Lu and Kyumin Lee and Xiaohu Liu and Chenlei Guo},
      year={2024},
      eprint={2403.06914},
      archivePrefix={arXiv},
      primaryClass={cs.CL},
      url={https://arxiv.org/abs/2403.06914}, 
}

@inproceedings{lin2004rouge,
    title = "{ROUGE}: A Package for Automatic Evaluation of Summaries",
    author = "Lin, Chin-Yew",
    booktitle = "Text Summarization Branches Out",
    month = jul,
    year = "2004",
    address = "Barcelona, Spain",
    publisher = "Association for Computational Linguistics",
    url = "https://aclanthology.org/W04-1013/",
    pages = "74--81"
}

@misc{liu2022peftbeatsicl,
      title={Few-Shot Parameter-Efficient Fine-Tuning is Better and Cheaper than In-Context Learning}, 
      author={Haokun Liu and Derek Tam and Mohammed Muqeeth and Jay Mohta and Tenghao Huang and Mohit Bansal and Colin Raffel},
      year={2022},
      eprint={2205.05638},
      archivePrefix={arXiv},
      primaryClass={cs.LG},
      url={https://arxiv.org/abs/2205.05638}, 
}

@misc{liu2023lostinthemiddle,
      title={Lost in the Middle: How Language Models Use Long Contexts}, 
      author={Nelson F. Liu and Kevin Lin and John Hewitt and Ashwin Paranjape and Michele Bevilacqua and Fabio Petroni and Percy Liang},
      year={2023},
      eprint={2307.03172},
      archivePrefix={arXiv},
      primaryClass={cs.CL},
      url={https://arxiv.org/abs/2307.03172}, 
}

@misc{liu2026shine,
      title={SHINE: A Scalable In-Context Hypernetwork for Mapping Context to LoRA in a Single Pass}, 
      author={Yewei Liu and Xiyuan Wang and Yansheng Mao and Yoav Gelbery and Haggai Maron and Muhan Zhang},
      year={2026},
      eprint={2602.06358},
      archivePrefix={arXiv},
      primaryClass={cs.CL},
      url={https://arxiv.org/abs/2602.06358}, 
}

@inproceedings{lv2024hyperlora,
    title = "{H}yper{L}o{RA}: Efficient Cross-task Generalization via Constrained Low-Rank Adapters Generation",
    author = "Lv, Chuancheng  and
      Li, Lei  and
      Zhang, Shitou  and
      Chen, Gang  and
      Qi, Fanchao  and
      Zhang, Ningyu  and
      Zheng, Hai-Tao",
    editor = "Al-Onaizan, Yaser  and
      Bansal, Mohit  and
      Chen, Yun-Nung",
    booktitle = "Findings of the Association for Computational Linguistics: EMNLP 2024",
    month = nov,
    year = "2024",
    address = "Miami, Florida, USA",
    publisher = "Association for Computational Linguistics",
    url = "https://aclanthology.org/2024.findings-emnlp.956/",
    doi = "10.18653/v1/2024.findings-emnlp.956",
    pages = "16376--16393"
}

@misc{mcmahan2017fl,
      title={Communication-Efficient Learning of Deep Networks from Decentralized Data}, 
      author={H. Brendan McMahan and Eider Moore and Daniel Ramage and Seth Hampson and Blaise Agüera y Arcas},
      year={2017},
      eprint={1602.05629},
      archivePrefix={arXiv},
      primaryClass={cs.LG},
      url={https://arxiv.org/abs/1602.05629}, 
}

@misc{mcmahan2018dpfl,
      title={Learning Differentially Private Recurrent Language Models}, 
      author={H. Brendan McMahan and Daniel Ramage and Kunal Talwar and Li Zhang},
      year={2018},
      eprint={1710.06963},
      archivePrefix={arXiv},
      primaryClass={cs.LG},
      url={https://arxiv.org/abs/1710.06963}, 
}

@misc{microsoft_phi_2025,
  author       = {{Microsoft Inc.}},
  title        = {Empowering innovation: The next generation of the phi family},
  howpublished = {Microsoft Azure Blog},
  year         = {2025},
  month        = {February},
  day          = {26},
  url          = {https://azure.microsoft.com/en-us/blog/empowering-innovation-the-next-generation-of-the-phi-family/},
  note         = {Accessed: 2026-04-21}
}

@misc{min2022metaicl,
      title={MetaICL: Learning to Learn In Context}, 
      author={Sewon Min and Mike Lewis and Luke Zettlemoyer and Hannaneh Hajishirzi},
      year={2022},
      eprint={2110.15943},
      archivePrefix={arXiv},
      primaryClass={cs.CL},
      url={https://arxiv.org/abs/2110.15943}, 
}

@misc{mosbach2023peft_vs_icl,
      title={Few-shot Fine-tuning vs. In-context Learning: A Fair Comparison and Evaluation}, 
      author={Marius Mosbach and Tiago Pimentel and Shauli Ravfogel and Dietrich Klakow and Yanai Elazar},
      year={2023},
      eprint={2305.16938},
      archivePrefix={arXiv},
      primaryClass={cs.CL},
      url={https://arxiv.org/abs/2305.16938}, 
}

@inproceedings{ni2019amazon,
    title = "Justifying Recommendations using Distantly-Labeled Reviews and Fine-Grained Aspects",
    author = "Ni, Jianmo  and
      Li, Jiacheng  and
      McAuley, Julian",
    editor = "Inui, Kentaro  and
      Jiang, Jing  and
      Ng, Vincent  and
      Wan, Xiaojun",
    booktitle = "Proceedings of the 2019 Conference on Empirical Methods in Natural Language Processing and the 9th International Joint Conference on Natural Language Processing (EMNLP-IJCNLP)",
    month = nov,
    year = "2019",
    address = "Hong Kong, China",
    publisher = "Association for Computational Linguistics",
    url = "https://aclanthology.org/D19-1018/",
    doi = "10.18653/v1/D19-1018",
    pages = "188--197"
}

@misc{ning2024userembedding,
      title={User-LLM: Efficient LLM Contextualization with User Embeddings}, 
      author={Lin Ning and Luyang Liu and Jiaxing Wu and Neo Wu and Devora Berlowitz and Sushant Prakash and Bradley Green and Shawn O'Banion and Jun Xie},
      year={2024},
      eprint={2402.13598},
      archivePrefix={arXiv},
      primaryClass={cs.CL},
      url={https://arxiv.org/abs/2402.13598}, 
}

@inproceedings{phang2023hypertuning,
  author       = {Jason Phang and
                  Yi Mao and
                  Pengcheng He and
                  Weizhu Chen},
  editor       = {Andreas Krause and
                  Emma Brunskill and
                  Kyunghyun Cho and
                  Barbara Engelhardt and
                  Sivan Sabato and
                  Jonathan Scarlett},
  title        = {HyperTuning: Toward Adapting Large Language Models without Back-propagation},
  booktitle    = {International Conference on Machine Learning, {ICML} 2023, 23-29 July
                  2023, Honolulu, Hawaii, {USA}},
  series       = {Proceedings of Machine Learning Research},
  volume       = {202},
  pages        = {27854--27875},
  publisher    = {{PMLR}},
  year         = {2023},
  _url          = {https://proceedings.mlr.press/v202/phang23a.html},
}

@misc{salemi2024lamp,
      title={LaMP: When Large Language Models Meet Personalization}, 
      author={Alireza Salemi and Sheshera Mysore and Michael Bendersky and Hamed Zamani},
      year={2024},
      eprint={2304.11406},
      archivePrefix={arXiv},
      primaryClass={cs.CL},
      url={https://arxiv.org/abs/2304.11406}, 
}

@article{schmidhuber1992fastweights,
  title={Learning to control fast-weight memories: An alternative to dynamic recurrent networks},
  author={Schmidhuber, J{\"u}rgen},
  journal={Neural Computation},
  volume={4},
  number={1},
  pages={131--139},
  year={1992},
  publisher={MIT Press One Rogers Street, Cambridge, MA 02142-1209, USA journals-info~…}
}

@misc{snell2022contextdistillation,
      title={Learning by Distilling Context}, 
      author={Charlie Snell and Dan Klein and Ruiqi Zhong},
      year={2022},
      eprint={2209.15189},
      archivePrefix={arXiv},
      primaryClass={cs.CL},
      url={https://arxiv.org/abs/2209.15189}, 
}

@misc{shamsian2021personalized_fl_hyp,
      title={Personalized Federated Learning using Hypernetworks}, 
      author={Aviv Shamsian and Aviv Navon and Ethan Fetaya and Gal Chechik},
      year={2021},
      eprint={2103.04628},
      archivePrefix={arXiv},
      primaryClass={cs.LG},
      url={https://arxiv.org/abs/2103.04628}, 
}

@misc{tan2025onepeftperuser,
      title={Democratizing Large Language Models via Personalized Parameter-Efficient Fine-tuning}, 
      author={Zhaoxuan Tan and Qingkai Zeng and Yijun Tian and Zheyuan Liu and Bing Yin and Meng Jiang},
      year={2025},
      eprint={2402.04401},
      archivePrefix={arXiv},
      primaryClass={cs.CL},
      url={https://arxiv.org/abs/2402.04401}, 
}

@inproceedings{tang2008citation,
author = {Tang, Jie and Zhang, Jing and Yao, Limin and Li, Juanzi and Zhang, Li and Su, Zhong},
title = {ArnetMiner: extraction and mining of academic social networks},
year = {2008},
isbn = {9781605581934},
publisher = {Association for Computing Machinery},
address = {New York, NY, USA},
url = {https://doi.org/10.1145/1401890.1402008},
doi = {10.1145/1401890.1402008},
booktitle = {Proceedings of the 14th ACM SIGKDD International Conference on Knowledge Discovery and Data Mining},
pages = {990–998},
numpages = {9},
location = {Las Vegas, Nevada, USA},
series = {KDD '08}
}

@inproceedings{vlske2017reddit,
  title={TL;DR: Mining Reddit to Learn Automatic Summarization},
  author={Michael V{\"o}lske and Martin Potthast and Shahbaz Syed and Benno Stein},
  booktitle={NFiS@EMNLP},
  year={2017},
  url={https://api.semanticscholar.org/CorpusID:2204603}
}

@misc{wei2022emergent,
      title={Emergent Abilities of Large Language Models}, 
      author={Jason Wei and Yi Tay and Rishi Bommasani and Colin Raffel and Barret Zoph and Sebastian Borgeaud and Dani Yogatama and Maarten Bosma and Denny Zhou and Donald Metzler and Ed H. Chi and Tatsunori Hashimoto and Oriol Vinyals and Percy Liang and Jeff Dean and William Fedus},
      year={2022},
      eprint={2206.07682},
      archivePrefix={arXiv},
      primaryClass={cs.CL},
      url={https://arxiv.org/abs/2206.07682}, 
}

@misc{workaccount22025contextrot, 
      title={Is there a half-life for the success rates of Ai Agents?: Hacker News}, 
      url={https://news.ycombinator.com/item?id=44308711#44310054}, 
      journal={Is there a half-life for the success rates of AI agents? | Hacker News}, 
      publisher={Hacker News}, 
      author={Workaccount2}, 
      year={2025}, 
      month={Jun},
}
\bibliographystyle{iclr2025_conference}

\newpage

\appendix

\section{Further Details on Hypernetwork Architecture and Training}
\label{app.hypernetwork}

\begin{table}
    \caption{Parameter counts of parts of the hypernetwork, for particular experiment configurations (LLM and rank of generated LoRA) applied in Section \ref{sec.exp}. The base LLM is the already-present on-device LLM. The additional parameters complete the hypernetwork.}
    \label{tab.parameter_counts}
    \begin{small}
        \begin{center}
        % \begin{sc}
        % \begingroup
        % \setlength{\tabcolsep}{4pt}                    
        \begin{tabular}{c|rrrr}
            \toprule
            \multirow{2}{*}{configuration:} & \textsc{Gemma3-1B-IT} & \textsc{Gemma3-1B-IT} & \textsc{Gemma1-2B-IT}  & \textsc{Gemma1-2B-IT} \\ 
             & \textsc{LoRA rank=4} & \textsc{LoRA rank=32} & \textsc{LoRA rank=4}  & \textsc{LoRA rank=32} \\ 
            \midrule[1.2pt]
            base LLM & 999885952 & 999885952 & 2506434560 & 2506434560 \\
            \midrule[1pt]
            embedder LoRA & \multirow{2}{*}{630272} & \multirow{2}{*}{630272} & \multirow{2}{*}{778752} & \multirow{2}{*}{778752} \\
            ($r=4$) & & & & \\
            embedding-to- & \multirow{3}{*}{13507072} & \multirow{3}{*}{87961088} & \multirow{3}{*}{16694784} & \multirow{3}{*}{108817920} \\
            matrix MLPs & & & & \\
            ($k=16$) & & & & \\
            total additional & 14137344 & 88591360 & 17473536 & 109596672 \\
            \midrule[1pt]
            generated LoRA & \multirow{3}{*}{630272} & \multirow{3}{*}{5009920} & \multirow{3}{*}{778752} & \multirow{3}{*}{6197760} \\
            (\textit{hypernetwork} & & & & \\
            \textit{output}) & & & & \\
            \bottomrule
        \end{tabular}
        % \endgroup
        % \end{sc}
        \end{center}
    \end{small}
    \vskip -0.2in
\end{table}

\begin{algorithm}
    \DontPrintSemicolon
    \SetKwInput{Input}{Input}
    %\SetAlgoLined
    \Input{LLM base params $\theta$; starting hypernetwork params $\psi^{(0)}$; users each with train dataset $\mathcal{D}_i$; hypernetwork function $\phi = H(x, \theta, \psi)$; loss function $l = L(x, \theta, \phi)$; total steps $T$; per-step user cohort size $I$; hypernetwork input count $N_H$; loss input batch size $N_L$; hypernetwork params optimizer $f$ and initial optimizer state $o^{(0)}$}
    % ~\\
    \For{{\it \bf step} $t \in \{1,\dots,T\}$ }{
      Sample a subset $\mathcal{S}^{(t)}$ of $I$ users;
      
      \For{{\it \bf user} $i \in \mathcal{S}^{(t)}$ {\it \bf in parallel}}{
        Sample (WOR) a batch $\mathcal{B}_H$ of $N_H$ examples and a batch $\mathcal{B}_L$ of $N_L$ examples from $\mathcal{D}_i$;
    
        Concatenate examples in $\mathcal{B}_H$ into single sequence: $x_H = [x_0, x_1, \cdots, x_{N_H - 1}]$\; 
        Generate user's LoRA: $\phi_i = H(x_H, \theta, \psi^{(t-1)})$\;
        Compute gradient $g_i$ of loss w.r.t. hypernetwork params: $g_i = \nabla_\psi L(\mathcal{B}_L, \theta, \phi_i)$\;
      }
      Compute average gradient: $g = \frac{1}{I} \sum_i g_i$\;
      Update hypernetwork params: $\psi^{(t)}, o^{(t)} = f (g, \psi^{(t-1)}, o^{(t-1)})$\;
    }
    % ~\\
    \Return $\psi^{(T)}$
    \caption{Hypernetwork training algorithm with end-to-end loss.}
    \label{algo.hyp_train}
\end{algorithm}

\paragraph{Pooling} We reduce the sequence representation vectors down to a single embedding vector via mean pooling, including only the sequence positions that were not padded in the input.

\section{Further Details on Datasets}
\label{app.datasets}

% TODO(saugenst): More details on the datasets described in subsec.datasets, and how we preprocessed data.

As discussed in Section \ref{sec.exp}, we used three personalization datasets: LongLaMP-3 and LongLaMP-4 from \citet{kumar2024longlamp} and LaMP-5 from \citet{salemi2024lamp}. 

Table \ref{tab.dataset_information} provides the processing parameters we used in configuring these datasets for experiments.

\begin{table}
    \caption{Information on datasets.}
    \label{tab.dataset_information}
    \begin{small}
        % \begin{center}
        % \begin{sc}
        % \begingroup
        \setlength{\tabcolsep}{4pt}                    
        \begin{tabular}{l|rr|rr|rrr}
            \toprule
            {~} & {prefix} & {target} & {\# train} & {\# examples} & {\# test} & {\# context examples} & {\# eval examples} \\
            {~} & {seq. len.} & {seq. len.} & {users} & {per train user} & {users} & {per test user} & {per test user} \\
            \midrule
            {LongLaMP-3} & 512 & 1024 & 14745 & 12 & 512 & 16 & 12\\
            \midrule
            {LongLaMP-4} & 256 & 1024 & 11442 & 12 & 640 & 16 & 12 \\
            \midrule
            {LaMP-5} & 512 & 64 & 9682 & 32 & 2496 & 16 & 16 \\
            \bottomrule
        \end{tabular}
        % \endgroup
        % \end{sc}
        % \end{center}
    \end{small}
    % \vskip -0.1in
\end{table}

When evaluating on the test set, each user's examples where partitioned into two disjoint sets. One set, consisting of 8 examples, was used as the user's `context', while the remaining examples where used for evaluation. To ensure fair and accurate comparisons, the context examples and evaluation examples were held consistent across all experiments and methods, i.e., so that the exact same context examples used in the input sequence with ICL were also used to form batches for fine-tuning with PEFT (and also used as the examples concatenated and fed to the hypernetwork).

\section{Hyperparameter Settings for Experiments}
\label{app.hyperparameters}

We provide more details on hyperparameter selections for the experiments discussed in Section \ref{sec.exp}.

In general, the computational resources used to train via these various methods consisted of clusters of accelerators, of sizes between 32 and 128 devices, each with 32GiB of memory. Experiments typically ran for several hours, with computation length depending in part on the dataset used (as sequence length is a major factor in processing time).

\subsection{\textsc{Non-personal baseline}}

A single common LoRA is trained for all users. For all models and datasets, we use a cohort size of 32 and a batch size of 4. At a given step, each user in the cohort computes a gradient with their batch, which was then averaged across the cohort, clipped, and used to calculate a parameter update via Adam. The learning rate was swept to achieve the lowest evaluation cross-entropy. For the experiments with \textsc{Gemma3-1B-IT}, the best learning rate was determined to be $1e-3$ with a cosine decay over 20000 steps. For the experiments with \textsc{Gemma1-2B-IT}, the best learning rate was determined to be $5e-4$ with a cosine decay over 20000 steps.

With this baseline, the context examples go unused, as no personalization is occurring.

\subsection{\textsc{ICL}}

This setup is largely analogous to the above, except for the fact that user-specific ICL prefixes were pre-appended, so that the LoRA is trained to be good at per-user ICL. For the experiments with \textsc{Gemma3-1B-IT}, the best learning rate was determined to be $1e-3$ with a cosine decay over 20000 steps. For the experiments with \textsc{Gemma1-2B-IT}, the best learning rate was determined to be $5e-4$ with a cosine decay over 20000 steps.

For each batch of data, the ICL prefix is constructed as follows. A limit of 3200 tokens is set. Candidate context examples are appended to an input sequence if they do not cause the total input sequence length to exceed this limit. The maximum number of context examples that will be appended is 8.

Note that we also undertook to test ICL with only the base LLM, i.e.~without any LoRA modifying the LLM's behavior. As discussed in Section \ref{sec.rw}, ICL is an emergent property that larger LLMs demonstrate very effectively but smaller LLMs struggle at (e.g. as conveyed in Figure 1.2 in \citet{brown2020icl}), so the expectation is that `pure' ICL would perform worse than `LoRA-modified ICL'. This is indeed confirmed by our experiments. See Tables \ref{tab.ll3_exps_summary}, \ref{tab.ll4_exps_summary}, and \ref{tab.lamp5_exps_summary}.

\subsection{\textsc{PEFT}}

For each user in the test set, their 8 context examples are formed into two batches of batch size 4, and then used to perform two steps of fine-tuning with the Adam optimizer. All users used a common learning rate (of 1e-4). We initialize the per-user optimizations from the checkpoint of the single common LoRA (i.e., the LoRA that we're using as our best non-personal baseline). 

Note that while the limited number of fine-tuning examples renders PEFT at somewhat of a disadvantage here, having limited amounts of previous examples is typically the norm in mobile AI applications, so this is faithful to the on-device setting that we focus this paper on.

\subsection{\textsc{Hypernetwork}}

We trained the hypernetwork with the Adam optimizer and learning rate of $3e-4$. As discussed in Section \ref{sec.train}, we used context distillation from the ICL LoRA in training, with a distillation temperature of 1.0 and multiplier of 1.0.

As mentioned in Section \ref{sec.arch}, the hypernetwork introduces a few additional hyperparameters, namely the rank of the embedder LoRA and the bottleneck width of the embedding-to-matrix MLPs. We uses a rank of 4 for the former, and a bottleneck dimension of 16 for the latter (for all experiments).

\section{Experiments: Detailed Results}
\label{app.results}

% TODO(saugenst): Other things to talk about
% - ICL sucks for LL-3.  Why?
% - Hypernetwork performance (at least for LL-3/LL-4) seems most responsive to increases in LoRA rank.

Tables \ref{tab.ll3_exps_summary}, \ref{tab.ll4_exps_summary}, and \ref{tab.lamp5_exps_summary} present the results of personalization experiments on the LongLaMP-3, LongLaMP-4, and LaMP-5 datasets, respectively. Bolded numbers indicate the best value in a column for a given experiment configuration (dataset and LoRA rank). In general, the hypernetwork-generated LoRAs exhibit the best ROUGE-1 score performance (or nearly so) in virtually all scenarios.

We also consider token cross-entropy on evaluation examples. Cross-entropy is more typically of interest in classification tasks, but we include it for completeness, and because it relates to the cross-entropy loss used at training time (and thus is indicative of generalization of training objectives).

\begin{table}
    \caption{LongLaMP-3 (Amazon review writing) experiments summary.}
    \label{tab.ll3_exps_summary}
    \begin{small}
        \begin{center}
        \begin{sc}
        \begingroup
        \setlength{\tabcolsep}{4pt}                    
        \begin{tabular}{cc|l|Sc|Sc}
            \toprule
            \multirow{4}{*}{~}
            & \multicolumn{2}{c|}{~} & \multicolumn{2}{c|}{ROUGE-1~($\uparrow$)} & \multicolumn{2}{c}{token x-entropy~($\downarrow$)} \\[3pt]
            & \multicolumn{2}{c|}{~} & {mean} & {\% of users} & {mean} & {\% of users} \\
            & \multicolumn{2}{c|}{~} & {over} & {improved vs.} & {over} & {improved vs.} \\
            & \multicolumn{2}{c|}{~} & {users} & {Single LoRA} & {users} & {Single LoRA} \\
            \midrule[1.0pt]
            \multirow{10}{*}{\rotatebox[origin=c]{90}{Gemma3-1B-IT}}
            & {no} & {Base LLM (non-personalized)} & 23.12 & \textendash & 3.566 & \textendash \\
            & {LoRA} & {Base LLM + \emph{per-user} ICL} & 23.34 & \textendash & 3.947 & \textendash \\
            \cmidrule[1.0pt]{2-7}
            & ~ & {Single LoRA (non-personalized)} & 33.06 & \textendash & 2.883 & \textendash \\
            & {LoRA} & {Single LoRA + \emph{per-user} ICL} & 32.57 & 37\% & 2.841 & 98\% \\
            & {rank=4} & {Fine-Tuned \emph{per-user} LoRAs} & 33.27 & 60\% & 2.878 & 100\% \\
            & ~ & {Hypernetwork \emph{per-user} LoRAs} & \textbf{34.08} & \textbf{77\%} & 2.885 & 40\% \\
            \cmidrule[1.0pt]{2-7}
            & ~ & {Single LoRA (non-personalized)} & 33.41 & \textendash & 2.833 & \textendash \\
            & {LoRA} & {Single LoRA + \emph{per-user} ICL} & 33.81 & 61\% & 2.786 & 99\% \\
            & {rank=32} & {Fine-Tuned \emph{per-user} LoRAs} & 33.51 & 56\% & 2.830 & 100\% \\
            & ~ & {Hypernetwork \emph{per-user} LoRAs} & \textbf{34.79} & \textbf{87\%} & 2.839 & 37\% \\
            \midrule[1.0pt]
            \multirow{10}{*}{\rotatebox[origin=c]{90}{Gemma1-2B-IT}}
            & {no} & {Base LLM (non-personalized)} & 22.13 & \textendash & 3.379 & \textendash \\
            & {LoRA} & {Base LLM + \emph{per-user} ICL} & 10.72 & \textendash & 3.243 & \textendash \\
            \cmidrule[1.0pt]{2-7}
            & ~ & {Single LoRA (non-personalized)} & 33.39 & \textendash & 2.685 & \textendash \\
            & {LoRA} & {Single LoRA + \emph{per-user} ICL} & 33.86 & 57\% & 2.638 & 99\% \\
            & {rank=4} & {Fine-Tuned \emph{per-user} LoRAs} & 33.59 & 57\% & 2.676 & 100\% \\
            & ~ & {Hypernetwork \emph{per-user} LoRAs} & \textbf{34.12} & \textbf{70\%} & 2.648 & 98\% \\
            \cmidrule[1.0pt]{2-7}
            & ~ & {Single LoRA (non-personalized)} & 33.67 & \textendash & 2.639 & \textendash \\
            & {LoRA} & {Single LoRA + \emph{per-user} ICL} & 34.23 & 61\% & 2.589 & 99\% \\
            & {rank=32} & {Fine-Tuned \emph{per-user} LoRAs} & 33.85 & 59\% & 2.632 & 99\% \\
            & ~ & {Hypernetwork \emph{per-user} LoRAs} & \textbf{34.58} & \textbf{77\%} & 2.616 & 91\% \\
            \bottomrule
        \end{tabular}
        \endgroup
        \end{sc}
        \end{center}
    \end{small}
    % \vskip -0.1in
\end{table}

\begin{table}
    \caption{LongLaMP-4 (Reddit post writing) experiments summary.}
    \label{tab.ll4_exps_summary}
    \begin{small}
        \begin{center}
        \begin{sc}
        \begingroup
        \setlength{\tabcolsep}{4pt}                    
        \begin{tabular}{cc|l|Sc|Sc}
            \toprule
            \multirow{4}{*}{~}
            & \multicolumn{2}{c|}{~} & \multicolumn{2}{c|}{ROUGE-1~($\uparrow$)} & \multicolumn{2}{c}{token x-entropy~($\downarrow$)} \\[3pt]
            & \multicolumn{2}{c|}{~} & {mean} & {\% of users} & {mean} & {\% of users} \\
            & \multicolumn{2}{c|}{~} & {over} & {improved vs.} & {over} & {improved vs.} \\
            & \multicolumn{2}{c|}{~} & {users} & {Single LoRA} & {users} & {Single LoRA} \\        
            \midrule[1.0pt]
            \multirow{10}{*}{\rotatebox[origin=c]{90}{Gemma3-1B-IT}}
            & {no} & {Base LLM (non-personalized)} & 22.91 & \textendash & 3.720 & \textendash \\
            & {LoRA} & {Base LLM + \emph{per-user} ICL} & 18.97 & \textendash & 4.134 & \textendash \\
            \cmidrule[1.0pt]{2-7}
            & ~ & {Single LoRA (non-personalized)} & 26.68 & \textendash & 3.031 & \textendash \\
            & {LoRA} & {Single LoRA + \emph{per-user} ICL} & 23.06 & 3\% & 3.001 & 99\% \\
            & {rank=4} & {Fine-Tuned \emph{per-user} LoRAs} & 26.84 & 60\% & 3.028 & 98\% \\
            & ~ & {Hypernetwork \emph{per-user} LoRAs} & \textbf{27.70} & \textbf{79\%} & 3.047 & 19\% \\
            \cmidrule[1.0pt]{2-7}
            & ~ & {Single LoRA (non-personalized)} & 26.86 & \textendash & 3.000 & \textendash \\
            & {LoRA} & {Single LoRA + \emph{per-user} ICL} & 26.66 & 41\% & 2.966 & 99\% \\
            & {rank=32} & {Fine-Tuned \emph{per-user} LoRAs} & 26.94 & 50\% & 2.999 & 94\% \\
            & ~ & {Hypernetwork \emph{per-user} LoRAs} & \textbf{28.23} & \textbf{83\%} & 3.034 & 5\% \\
            \midrule[1.0pt]
            \multirow{10}{*}{\rotatebox[origin=c]{90}{Gemma1-2B-IT}}
            & {no} & {Base LLM (non-personalized)} & 20.10 & \textendash & 3.736 & \textendash \\
            & {LoRA} & {Base LLM + \emph{per-user} ICL} & 11.70 & \textendash & 3.481 & \textendash \\
            \cmidrule[1.0pt]{2-7}
            & ~ & {Single LoRA (non-personalized)} & 27.08 & \textendash & 2.861 & \textendash \\
            & {LoRA} & {Single LoRA + \emph{per-user} ICL} & 24.96 & 12\% & 2.826 & 99\% \\
            & {rank=4} & {Fine-Tuned \emph{per-user} LoRAs} & 27.28 & 59\% & 2.855 & 99\% \\
            & ~ & {Hypernetwork \emph{per-user} LoRAs} & \textbf{27.84} & \textbf{76\%} & 2.839 & 96\% \\
            \cmidrule[1.0pt]{2-7}
            & ~ & {Single LoRA (non-personalized)} & 27.37 & \textendash & 2.824 & \textendash \\
            & {LoRA} & {Single LoRA + \emph{per-user} ICL} & 27.41 & 50\% & 2.785 & 99\% \\
            & {rank=32} & {Fine-Tuned \emph{per-user} LoRAs} & 27.53 & 59\% & 2.819 & 100\% \\
            & ~ & {Hypernetwork \emph{per-user} LoRAs} & \textbf{28.13} & \textbf{79\%} & 2.844 & 11\% \\
            \bottomrule
        \end{tabular}
        \endgroup
        \end{sc}
        \end{center}
    \end{small}
    % \vskip -0.1in
\end{table}

\begin{table}
    \caption{LaMP-5 (scholarly title writing) experiments summary.}
    \label{tab.lamp5_exps_summary}
    \begin{small}
        \begin{center}
        \begin{sc}
        \begingroup
        \setlength{\tabcolsep}{4pt}                    
        \begin{tabular}{cc|l|Sc|Sc}
            \toprule
            \multirow{4}{*}{~}
            & \multicolumn{2}{c|}{~} & \multicolumn{2}{c|}{ROUGE-1~($\uparrow$)} & \multicolumn{2}{c}{token x-entropy~($\downarrow$)} \\[3pt]
            & \multicolumn{2}{c|}{~} & {mean} & {\% of users} & {mean} & {\% of users} \\
            & \multicolumn{2}{c|}{~} & {over} & {improved vs.} & {over} & {improved vs.} \\
            & \multicolumn{2}{c|}{~} & {users} & {Single LoRA} & {users} & {Single LoRA} \\        
            \midrule[1.0pt]
            \multirow{10}{*}{\rotatebox[origin=c]{90}{Gemma3-1B-IT}}
            & {no} & {Base LLM (non-personalized)} & 5.43 & \textendash & 6.293 & \textendash \\
            & {LoRA} & {Base LLM + \emph{per-user} ICL} & 6.00 & \textendash & 6.944 & \textendash \\
            \cmidrule[1.0pt]{2-7}
            & ~ & {Single LoRA (non-personalized)} & 41.23 & \textendash & 1.952 & \textendash \\
            & {LoRA} & {Single LoRA + \emph{per-user} ICL} & 41.03 & 47\% & 1.907 & 75\% \\
            & {rank=4} & {Fine-Tuned \emph{per-user} LoRAs} & 41.29 & 50\% & 1.949 & 69\% \\
            & ~ & {Hypernetwork \emph{per-user} LoRAs} & \textbf{41.74} & \textbf{55\%} & 1.906 & 81\% \\
            \cmidrule[1.0pt]{2-7}
            & ~ & {Single LoRA (non-personalized)} & 41.85 & \textendash & 1.900 & \textendash \\
            & {LoRA} & {Single LoRA + \emph{per-user} ICL} & 41.69 & 48\% & 1.850 & 75\% \\
            & {rank=32} & {Fine-Tuned \emph{per-user} LoRAs} & 41.91 & 49\% & 1.897 & 66\% \\
            & ~ & {Hypernetwork \emph{per-user} LoRAs} & \textbf{42.29} & \textbf{53\%} & 1.844 & 73\% \\
            \midrule[1.0pt]
            \multirow{10}{*}{\rotatebox[origin=c]{90}{Gemma1-2B-IT}}
            & {no} & {Base LLM (non-personalized)} & 7.57 & \textendash & 3.596 & \textendash \\
            & {LoRA} & {Base LLM + \emph{per-user} ICL} & 3.62 & \textendash & 3.811 & \textendash \\
            \cmidrule[1.0pt]{2-7}
            & ~ & {Single LoRA (non-personalized)} & 43.46 & \textendash & 1.791 & \textendash \\
            & {LoRA} & {Single LoRA + \emph{per-user} ICL} & 42.69 & 43\% & 1.750 & 76\% \\
            & {rank=4} & {Fine-Tuned \emph{per-user} LoRAs} & 43.53 & 50\% & 1.787 & 4\% \\
            & ~ & {Hypernetwork \emph{per-user} LoRAs} & \textbf{43.93} & \textbf{55\%} & 1.757 & 70\% \\
            \cmidrule[1.0pt]{2-7}
            & ~ & {Single LoRA (non-personalized)} & 44.11 & \textendash & 1.758 & \textendash \\
            & {LoRA} & {Single LoRA + \emph{per-user} ICL} & 43.38 & 43\% & 1.713 & 77\% \\
            & {rank=32} & {Fine-Tuned \emph{per-user} LoRAs} & \textbf{44.18} & 49\% & 1.753 & 72\% \\
            & ~ & {Hypernetwork \emph{per-user} LoRAs} & 44.16 & \textbf{51\%} & 1.712 & 60\% \\
            \bottomrule
        \end{tabular}
        \endgroup
        \end{sc}
        \end{center}
    \end{small}
    % \vskip -0.1in
\end{table}

\section{Experiments: Ablations}
\label{app.ablations}

\subsection{Regularization vs. Distillation}

As mentioned in Section \ref{sec.train}, training the hypernetwork with distillation from an ICL-trained LoRA is beneficial to hypernetwork training. We also experimented with an alternative training modification: applying $L_1$-regularization to the values at the bottleneck of the embedding-to-matrix MLPs. Recall from Section \ref{sec.arch} the interpretation that the values of this bottleneck represent the coefficients of `principal' matrices that will be linearly combined to form output LoRA matrices. Ideally, we'd like the hypernetwork to be rewarded for involving the fewest principal matrices (to encourage learning the most relevant matrices). Unfortunately, the $L_0$-regularization that would achieve this is non-differentiable, so we instead apply something similar (but differentiable) in the form of $L_1$-regularization.

Tables \ref{tab.ll3_exps_regularization} and \ref{tab.ll4_exps_regularization} show the performance of this regularized version (on LongLaMP-3 and LongLaMP-4, respectively) alongside the other personalization methods. As can be seen, while the distillation-trained hypernetwork still generally achieves the best ROUGE-1 performance, the regularization-trained is close. It also beats the other personalization methods at ROUGE-1 (both in mean over users as well as percentage of users improved). Interestingly, it performs much better at cross-entropy then the distillation-trained hypernetwork.

\begin{table}
    \caption{LongLaMP-3, Hypernetworks trained via distillation vs. via regularization.}
    \label{tab.ll3_exps_regularization}
    \begin{small}
        \begin{center}
        \begin{sc}
        \begingroup
        \setlength{\tabcolsep}{4pt}                    
        \begin{tabular}{cc|l|Sc|Sc}
            \toprule
            \multirow{4}{*}{~}
            & \multicolumn{2}{c|}{~} & \multicolumn{2}{c|}{ROUGE-1~($\uparrow$)} & \multicolumn{2}{c}{token x-entropy~($\downarrow$)} \\[3pt]
            & \multicolumn{2}{c|}{~} & {mean} & {\% of users} & {mean} & {\% of users} \\
            & \multicolumn{2}{c|}{~} & {over} & {improved vs.} & {over} & {improved vs.} \\
            & \multicolumn{2}{c|}{~} & {users} & {Single LoRA} & {users} & {Single LoRA} \\
            \midrule[1.0pt]
            \multirow{10}{*}{\rotatebox[origin=c]{90}{Gemma3-1B-IT}}
            & ~ & {Single LoRA (non-personalized)} & 33.06 & \textendash & 2.883 & \textendash \\
            & {LoRA} & {Single LoRA + \emph{per-user} ICL} & 32.57 & 37\% & 2.841 & 98\% \\
            & {rank=4} & {Fine-Tuned \emph{per-user} LoRAs} & 33.27 & 60\% & 2.878 & 100\% \\
            & ~ & {Hypernetwork (\emph{via distillation})} & \textbf{34.08} & 77\% & 2.885 & 40\% \\
            & ~ & {Hypernetwork (\emph{via regularization})} & 34.03 & \textbf{78\%} & 2.853 & 98\% \\
            \cmidrule[1.0pt]{2-7}
            & ~ & {Single LoRA (non-personalized)} & 33.41 & \textendash & 2.833 & \textendash \\
            & {LoRA} & {Single LoRA + \emph{per-user} ICL} & 33.81 & 61\% & 2.786 & 99\% \\
            & {rank=32} & {Fine-Tuned \emph{per-user} LoRAs} & 33.51 & 56\% & 2.830 & 100\% \\
            & ~ & {Hypernetwork (\emph{via distillation})} & \textbf{34.79} & \textbf{87\%} & 2.839 & 37\% \\
            & ~ & {Hypernetwork (\emph{via regularization})} & 34.28 & 74\% & 2.815 & 90\% \\
            \bottomrule
        \end{tabular}
        \endgroup
        \end{sc}
        \end{center}
    \end{small}
    \vskip -0.1in
\end{table}

\begin{table}
    \caption{LongLaMP-4, Hypernetworks trained via distillation vs. via regularization.}
    \label{tab.ll4_exps_regularization}
    \begin{small}
        \begin{center}
        \begin{sc}
        \begingroup
        \setlength{\tabcolsep}{4pt}                    
        \begin{tabular}{cc|l|Sc|Sc}
            \toprule
            \multirow{4}{*}{~}
            & \multicolumn{2}{c|}{~} & \multicolumn{2}{c|}{ROUGE-1~($\uparrow$)} & \multicolumn{2}{c}{token x-entropy~($\downarrow$)} \\[3pt]
            & \multicolumn{2}{c|}{~} & {mean} & {\% of users} & {mean} & {\% of users} \\
            & \multicolumn{2}{c|}{~} & {over} & {improved vs.} & {over} & {improved vs.} \\
            & \multicolumn{2}{c|}{~} & {users} & {Single LoRA} & {users} & {Single LoRA} \\        
            \midrule[1.0pt]
            \multirow{10}{*}{\rotatebox[origin=c]{90}{Gemma3-1B-IT}}
            & ~ & {Single LoRA (non-personalized)} & 26.68 & \textendash & 3.031 & \textendash \\
            & {LoRA} & {Single LoRA + \emph{per-user} ICL} & 23.06 & 3\% & 3.001 & 99\% \\
            & {rank=4} & {Fine-Tuned \emph{per-user} LoRAs} & 26.84 & 60\% & 3.028 & 98\% \\
            & ~ & {Hypernetwork (\emph{via distillation})} & \textbf{27.70} & \textbf{79\%} & 3.047 & 19\% \\
            & ~ & {Hypernetwork (\emph{via regularization})} & 27.10 & 65\% & 3.011 & 96\% \\
            \cmidrule[1.0pt]{2-7}
            & ~ & {Single LoRA (non-personalized)} & 26.86 & \textendash & 3.000 & \textendash \\
            & {LoRA} & {Single LoRA + \emph{per-user} ICL} & 26.66 & 41\% & 2.966 & 99\% \\
            & {rank=32} & {Fine-Tuned \emph{per-user} LoRAs} & 26.94 & 50\% & 2.999 & 94\% \\
            & ~ & {Hypernetwork (\emph{via distillation})} & \textbf{28.23} & \textbf{83\%} & 3.034 & 5\% \\
            & ~ & {Hypernetwork (\emph{via regularization})} & 27.70 & 73\% & 2.985 & 91\% \\
            \bottomrule
        \end{tabular}
        \endgroup
        \end{sc}
        \end{center}
    \end{small}
    \vskip -0.1in
\end{table}

\subsection{Mismatching User Contexts}

As a general check that the hypernetwork is actually making use of users' contexts, we ran an experiment where at test time we mismatched the data so that every user had their own evaluation examples, but had the context examples of a different user. If the hypernetwork is accurately `locking on to signal' in the context, then this should result in every user receiving a LoRA that is not well-suited to them, and the ROUGE-1 scores should drop noticeably. We ran this experiment on LongLaMP-3 with \textsc{Gemma3-1B-IT} (Table \ref{tab.ll3_exps_mismatch}), and this indeed exactly what we see. When contexts get mismatched, the personalization quality drops below all other methods (including the non-personalized baseline).

\begin{table}
    \caption{LongLaMP-3, Hypernetworks with mismatching user contexts.}
    \label{tab.ll3_exps_mismatch}
    \begin{small}
        \begin{center}
        \begin{sc}
        \begingroup
        \setlength{\tabcolsep}{4pt}                    
        \begin{tabular}{cc|l|Sc|Sc}
            \toprule
            \multirow{4}{*}{~}
            & \multicolumn{2}{c|}{~} & \multicolumn{2}{c|}{ROUGE-1~($\uparrow$)} & \multicolumn{2}{c}{token x-entropy~($\downarrow$)} \\[3pt]
            & \multicolumn{2}{c|}{~} & {mean} & {\% of users} & {mean} & {\% of users} \\
            & \multicolumn{2}{c|}{~} & {over} & {improved vs.} & {over} & {improved vs.} \\
            & \multicolumn{2}{c|}{~} & {users} & {Single LoRA} & {users} & {Single LoRA} \\
            \midrule[1.0pt]
            \multirow{10}{*}{\rotatebox[origin=c]{90}{Gemma3-1B-IT}}
            & ~ & {Single LoRA (non-personalized)} & 33.06 & \textendash & 2.883 & \textendash \\
            & {LoRA} & {Single LoRA + \emph{per-user} ICL} & 32.57 & 37\% & 2.841 & 98\% \\
            & {rank=4} & {Fine-Tuned \emph{per-user} LoRAs} & 33.27 & 60\% & 2.878 & 100\% \\
            & ~ & {Hypernetwork \emph{per-user} LoRAs} & \textbf{34.08} & \textbf{77\%} & 2.885 & 40\% \\
            & ~ & {Hypernetwork \emph{mismatched contexts}} & 31.93 & 26\% & 2.974 & 0\% \\
            \cmidrule[1.0pt]{2-7}
            & ~ & {Single LoRA (non-personalized)} & 33.41 & \textendash & 2.833 & \textendash \\
            & {LoRA} & {Single LoRA + \emph{per-user} ICL} & 33.81 & 61\% & 2.786 & 99\% \\
            & {rank=32} & {Fine-Tuned \emph{per-user} LoRAs} & 33.51 & 56\% & 2.830 & 100\% \\
            & ~ & {Hypernetwork \emph{per-user} LoRAs} & \textbf{34.79} & \textbf{87\%} & 2.839 & 37\% \\
            & ~ & {Hypernetwork \emph{mismatched contexts}} & 33.10 & 43\% & 2.943 & 0\% \\
            \bottomrule
        \end{tabular}
        \endgroup
        \end{sc}
        \end{center}
    \end{small}
    \vskip -0.1in
\end{table}

\end{document}